\documentclass[sigconf]{acmart}  

\AtBeginDocument{%
  \providecommand\BibTeX{{%
    \normalfont B\kern-0.5em{\scshape i\kern-0.25em b}\kern-0.8em\TeX}}}

\setcopyright{acmcopyright}
\copyrightyear{2023}
\acmYear{2023}
\acmDOI{XXXXXXX.XXXXXXX}

\acmConference[SIGMOD '23]{In Proceedings of the 2023 International Conference on Management of Data}{June 18--23, 2023}{Seattle, WA, USA}
\acmPrice{15.00}
\acmISBN{978-1-4503-XXXX-X/18/06}

\usepackage{algorithm}
\usepackage{algpseudocode}
\usepackage{graphicx}
\usepackage{textcomp}
\usepackage{xcolor, colortbl}
\usepackage{multirow}
\usepackage{comment}
\usepackage{subfigure}
\usepackage{xfrac}
\usepackage{enumitem}
\usepackage{caption}
\usepackage{amsmath,amsfonts,bm}

\def\1{\bm{1}}

\def\vzero{{\bm{0}}}

\def\vbeta{{\bm{\beta}}}
\def\vepsilon{{\bm{\epsilon}}}
\def\vlambda{{\bm{\lambda}}}
\def\vsigma{{\bm{\sigma}}}
\def\vzeta{{\bm{\zeta}}}

\def\vb{{\bm{b}}}
\def\vc{{\bm{c}}}
\def\vd{{\bm{d}}}

\def\vh{{\bm{h}}}

\def\vj{{\bm{j}}}

\def\vr{{\bm{r}}}

\def\vx{{\bm{x}}}
\def\vy{{\bm{y}}}

\def\mI{{\bm{I}}}
\def\mJ{{\bm{J}}}

\def\mQ{{\bm{Q}}}

\def\mW{{\bm{W}}}
\def\mX{{\bm{X}}}

\def\mZ{{\bm{Z}}}

\DeclareMathAlphabet{\mathsfit}{\encodingdefault}{\sfdefault}{m}{sl}
\SetMathAlphabet{\mathsfit}{bold}{\encodingdefault}{\sfdefault}{bx}{n}

\def\gN{{\mathcal{N}}}
\def\gO{{\mathcal{O}}}

\def\sR{{\mathbb{R}}}

\newcommand{\Var}{\mathrm{Var}}

\newcommand{\Cov}{\mathrm{Cov}}

\DeclareMathOperator*{\argmin}{arg\,min}

\DeclareMathOperator{\tr}{tr}
\DeclareMathOperator{\diag}{diag}
\DeclareMathOperator{\offdiag}{off-diag}

\newcommand{\yh}[1]{\textcolor{black}{#1}}

\makeatletter
\newcommand{\clearsubcaptcounter}{\setcounter{sub\@captype}{0}}
\makeatother

\allowdisplaybreaks[3]
\begin{document}

\title[BALANCE: Bayesian Linear Attribution for Root Cause Localization]{BALANCE: Bayesian Linear Attribution for \\ Root Cause Localization}
%

\author{Chaoyu Chen$^\dag$}
\affiliation{%
  \institution{Ant Group}
  \country{China}}
\email{chris.ccy@antgroup.com}

\author{Hang Yu$^\dag$}
\affiliation{%
  \institution{Ant Group}
  \country{China}}
\email{hyu.hugo@antgroup.com}

\author{Zhichao Lei}
\affiliation{%
  \institution{Ant Group}
  \country{China}}
\email{leizhichao.lzc@antgroup.com}

\author{Jianguo Li}
\authornote{Corresponding author. $^\dag$ Two authors contributed equally to this work.}
\affiliation{%
  \institution{Ant Group}
  \country{China}}
\email{lijg.zero@antgroup.com}

\author{Shaokang Ren}
\affiliation{%
  \institution{Ant Group}
  \country{China}}
\email{renshaokang.rsk@antgroup.com}

\author{Tingkai Zhang}
\affiliation{%
  \institution{Ant Group}
  \country{China}}
\email{tingkai.ztk@antgroup.com}

\author{Silin Hu}
\affiliation{%
  \institution{Ant Group}
  \country{China}}
\email{husilin.hsl@antgroup.com}

\author{Jianchao Wang}
\affiliation{%
  \institution{Ant Group}
  \country{China}}
\email{luli.wjc@antgroup.com}

\author{Wenhui Shi}
\affiliation{%
  \institution{Ocean Base}
  \country{China}}
\email{yushun.swh@oceanbase.com}
\renewcommand{\shortauthors}{Chen et al.}

\begin{abstract}
Root Cause Analysis (RCA) plays an indispensable role in distributed data system maintenance and operations, as it bridges the gap between fault detection and system recovery. Existing works mainly study multidimensional localization or graph-based root cause localization. This paper opens up the possibilities of exploiting the recently developed framework of explainable AI (XAI) for the purpose of RCA. In particular, we propose BALANCE (BAyesian Linear AttributioN for root CausE localization), which formulates the problem of RCA through the lens of attribution in XAI and seeks to explain the anomalies in the target KPIs by the behavior of the candidate root causes. BALANCE consists of three innovative components. First, we propose a Bayesian multicollinear feature selection (BMFS) model to predict the target KPIs given the candidate root causes in a forward manner while promoting sparsity and concurrently paying attention to the correlation between the candidate root causes. Second, we introduce attribution analysis to compute the attribution score for each candidate in a backward manner. Third, we merge the estimated root causes related to each KPI if there are multiple KPIs. \yh{We extensively evaluate the proposed BALANCE method on one synthesis dataset as well as three real-world RCA tasks, that is, bad SQL localization, container fault localization, and fault type diagnosis for Exathlon.} Results show that BALANCE outperforms the state-of-the-art (SOTA) methods in terms of accuracy with the least amount of running time, \yh{and achieves at least $6\%$ notably higher accuracy than SOTA methods for real tasks.} BALANCE has been deployed to production to tackle real-world RCA problems, and the online results further advocate its usage for real-time diagnosis in distributed data systems.

\end{abstract}



\maketitle

\section{Introduction}\label{sec:intro}

System faults and incidents have a possibly tremendous influence on distributed data systems which are widely adopted in modern information technology (IT) and financial companies, since they may lead to system outrage and further incur astounding financial loss and jeopardize customer trust~\cite{nair2015learning}. It has been reported by Forbes that every year IT downtime costs an estimated \$26.5 billion in lost revenue alone, not to mention the indirect expense, including lost customers and references. Thus, it is imperative to conduct fast and precise fault diagnosis and recovery before they become service-impacting. A central task in fault diagnosis and recovery is root cause analysis (RCA), which bridges the gap between fault detection and recovery~\cite{dang2019aiops, jeyakumar2019explainit}. 

Currently, the task of RCA is mainly accomplished by site reliability engineers (SREs) with rich operation experience. Unfortunately, such manual work becomes prohibitively slow due to the increase of the scale and complexity of the architecture as well as the dynamic and unpredictable nature of the system metrics and events, thus deviating from the requirement of efficiency. \yh{Indeed, as mentioned in~\cite{ma2020diagnosing}, it can take as long as several hours of manual work to diagnose the root causes of intermittent slow queries in distributed database systems.} This has sparked considerable research efforts toward designing automated RCA algorithms based on machine learning so as to provide aid in saving time and ultimately money.

Literature on RCA algorithms can be broadly divided into two categories. The first one focuses on multidimensional root cause localization~\cite{bhagwan2014adtributor, sun2018hotspot, zhang2021halo}, which seeks to explain the abnormal behavior of the additive key performance indicators (KPIs) by identifying the fault-indicating combinations of their corresponding multi-dimensional attributes. The success of these algorithms relies on two assumptions: 1) the value of the KPI in each dimension equals the sum of the values of its attributes and 2) all the KPIs and their attributes can be monitored. However, these two assumptions can be too restrictive in real-world problems, and a more practical setting is to attribute the anomalies to root cause candidates without additive assumptions while allowing for missing data. On the other hand, the second category revolves around graph-based RCA algorithms~\cite{kim2013root, wang2018cloudranger, wang2019grano, qiu2020causality}. These approaches typically first construct a causal graph based on tracing service calls or causal discovery algorithms~\cite{zhang2021fast} and then find the root cause node via rule-based traversing or random walk. A major impediment to the application of tracing graphs and rule-based traversing is that it is system invasive and typically incurs arduous work on enumerating all traces and rules. As an alternative, causal discovery methods are employed to learn the graph structure as in~\cite{wang2018cloudranger}. Unfortunately, the causal discovery methods suffer from both high computational and sample complexity~\cite{zhang2021fast}, and in consequence, they can be distressingly slow for large graphs and may lead to inaccurate results when the number of observations for all metrics in the graph is small. After obtaining the graph, the random walk methods are heuristic and might fail to converge to the root cause when the number of random walks is not sufficiently large.

In this paper, we explore alternatives and recast the RCA problem as a feature attribution problem~\cite{linardatos2020explainable}. To the best of our knowledge, we are among the first to analyze the root cause through the lens of attribution. As a commonly used tool in explainable AI (XAI), attribution methods assign attribution scores to input features, the absolute value of which represents their importance to the model prediction or performance~\cite{linardatos2020explainable}. Analogously, we aim to find the root causes that can best explain the alarmed KPIs in RCA problems. The attribution scores of the candidate causes represent their relevance or contribution to the alarmed KPIs. \yh{As a motivating example, in database systems, ``bad SQLs'' is referred to as SQLs with deteriorated performance due to indexing errors or changes in the execution plan. The performance deterioration of these SQLs typically leads to anomalies in the tenant KPIs and may severely influence the user experience. In this case, the target ($\vy$) are the tenant KPIs and the candidate causes ($\mX$) are SQL metrics.} 

An attribution task can then be accomplished in two steps: first, a forward model is constructed that exploits the input features (i.e., candidate causes) to predict the outputs (i.e., alarmed KPIs), and next, the significance of the input features are evaluated through attribution approaches in a backward manner. \yh{Particularly in the bad SQL localization example, the number of candidate SQLs $p$ varies in each case and can be as large as thousands, whereas the number of observations $n$ (the length of the corresponding time series) is typically small since we only focus on the part around the anomalies. In other words, the dimension $p$ can be larger than the sample size $n$ in the RCA problems. To address this issue and to automate the feature selection process, we adopt sparse linear models as the forward model due to their high flexibility, efficiency, and interpretability.} Furthermore, the candidate causes are usually correlated with each other, and there often exist missing values. To tackle these problems that plague linear models, we propose a novel Bayesian multicollinear feature selection (BMFS) model. Afterward, we provide the attribution score for each candidate cause from different perspectives, including sensitivity and salience. Finally, we merge results when there exist multiple alarmed KPIs and each of them is attributed to a different set of root causes. We name the overall model BALANCE (BAyesian Linear AttributioN for root CausE localization). 

We would like to point out that both the multidimensional RCA and the graph-based RCA can be formulated from the perspective of attribution. Specifically, we can regard the multidimensional RCA as attributing the anomalies in the KPIs to the combinations of their attributes. It follows that the additive constraints in the multidimensional RCA can be removed, and hence, we only need to consider the abnormal attributes under this scenario. On the other hand, by regarding all abnormal nodes in the graph as candidate causes, BALANCE can be used to identify the root cause efficiently even though the graph structure is not available or cannot be reliably learned, which is often the case in practice. Viewed another way, BALANCE can also be used as a building block to construct causal graphs, since linear regression models are frequently used for causal discovery~\cite{zheng2018dags}. Given the graph, BALANCE serves as a better substitute for random walks as it does not require a large number of random walks and so is more efficient. 

We validate the usefulness of BALANCE on four datasets. First, we generate synthetic data with a different number of input features, different levels of multicollinearity, noise, and sparsity, and different proportions of missing values, and then compare various forward models including the proposed BMFS, Lasso, E-Net (Elastic net), and ARD (Automatic Relevance Determination). We find that BMFS typically recovers the underlying true regression coefficients the best with comparable or even shorter running time, especially when there exists multicollinearity among the input features. Furthermore, we utilize the proposed method to address three real-world RCA problems. In the first problem, \yh{we deal with the problem of bad SQL localization as mentioned before}. Our results show that the proposed method can identify the human-labeled root cause SQLs in fewer than 2 seconds per case with accuracy as high as $83.3\%$, whereas it takes 3 minutes for SREs on average. The second application copes with the problem of container fault localization, whose objective is to attribute the abnormal trace failures in a container to the metrics of the container, such as CPU usage, memory usage, TCP, etc, and facilitate the self-healing process. The proposed method can achieve an $F_1$-score of $0.86$, which is at least $20\%$ higher than other baseline methods. \yh{Finally, we apply BALANCE to a public dataset, Exathlon~\cite{jacob2021exathlon}, for the purpose of fault type diagnosis, and the resulting accuracy is again $6\%$ higher than the SOTA methods.} Note that the first application handles KPIs and candidates that are homogeneous while the latter two tackles heterogeneous ones. The advantageous performance of BALANCE in all three scenes shows that attribution-based RCA can be an effective and efficient tool for general and practical RCA cases.

In summary, our contribution includes:
\begin{itemize}[leftmargin=1.5em,itemsep= 2pt,topsep = 2pt,partopsep=2pt]
    \item To the best of our knowledge, we are among the first to formulate the RCA problem from the perspective of attribution analysis developed in the field of XAI. To be specific, we explain the anomaly in the target KPI by attributing it to the behavior of the candidate's root causes.
    \item We propose a novel forward model BMFS that can automatically select relevant candidates while taking their correlations into account at the same time.
    \item We apply the proposed BALANCE approach to three real-world problems, including bad SQL localization, container fault localization, and fault type diagnosis for Exathlon. All three scenes demonstrate the effectiveness of BALANCE. We further deploy BALANCE to production for the former two applications.
\end{itemize}

\vspace{-1.5ex}
\section{Related Works}
Since the proposed model is related to RCA, attribution, and sparse linear models, we provide a brief review of each of them below.

\vspace{-1.5ex}
\subsection{Root Cause Localization}

As aforementioned, there are broadly two strategies for root cause localization. The first one considers the multidimensional root cause localization methods, such as Adtributor~\cite{bhagwan2014adtributor}, Hotspot~\cite{sun2018hotspot}, and HALO~\cite{zhang2021halo}. More concretely, Adtributor~\cite{bhagwan2014adtributor} finds the root cause in each dimension by selecting the most abnormal dimension values. However, Adtributor assumes that the root cause lies in one dimension, which can be too restrictive in practice. To extend Adtributor to the case of multidimensional root causes, Hotspot~\cite{sun2018hotspot} propagates the anomaly of the KPIs to different dimensions via the ripple effect and further defines the attribution scores of different dimension combinations by replacing their real values with the predicted ones and further computing the differences with and without the replacement. Unfortunately, the consequent search space is typically very large. Although Hotspot employs Monte Carlo tree search and a hierarchical pruning strategy to reduce the computational cost, it can still be too slow for large-scale practical problems. Another drawback of both Adtributor and Hotspot is that they fail to consider the possible dependency among dimensions. As a solution, HALO~\cite{zhang2021halo} learns the hierarchical dependency structure of the dimensions via conditional entropy and then looks for the root cause by traversing in this structure. These kinds of methods can be viewed as a special case of attribution methods since they try to attribute the anomalies in the KPIs to the combinations of the associated multidimensional attributes. Note that the target KPIs in this case is the sum of the attribute values along each dimension. However, in a more general setting, the target KPIs may be influenced by the root cause candidates in a non-additive manner. The proposed BALANCE approach provides a recipe for this problem.

The second strategy focuses on the graph-based causal inference~ \cite{soldani2022anomaly}. \yh{MonitorRank~\cite{kim2013root} and Microhecl~\cite{liu2021microhecl} firstly introduce service level RCA on known services chain architecture given by the distributed tracing system. On the other hand, for applications without known graphs, CauseInfer~\cite{chen2014causeinfer} and CloudRanger~\cite{wang2018cloudranger} build a causal graph using the PC Algorithm~\cite{spirtes1991algorithm}. Once the graph is obtained, statistical root causes are typically inferred via personalized page rank~\cite{weng2018root}, breadth-first search~\cite{chen2014causeinfer,liu2021microhecl}, random walk~\cite{kim2013root,wang2018cloudranger}, etc.} To make the graph more reliable, OM Graph~\cite{qiu2020causality} considers the prior knowledge from a knowledge graph with entities representing all software and hardware in a distributed data system, during the construction of the causal graph. Furthermore, CloudRCA~\cite{zhang2021cloudrca} optimizes OM Graph by building graphs based on multiple sources of data, including monitoring metrics, logs, and expert knowledge. Note that the PC algorithm is still used in both OM graph and CloudRCA for graph construction. As pointed out in $\S$\ref{sec:intro}, it may be resource-consuming to construct graphs based on tracing service calls or other sources of prior knowledge, due to the large-scale and complicated nature of the entire architecture, while it is unreliable to build graphs utilizing causal discovery algorithms (e.g., the PC algorithm) given the limited length of the time series for the metrics during the anomalies. Such shortcomings hamper the practice use of graph-based RCA. \yh{Another line of research seeks to construct the causal graph by identifying the lagged temporal dependence between different metrics~\cite{moraffah2021causal,assaad2022survey}. For instance, Granger causality is adopted in~\cite{thalheim2017sieve, aggarwal2020localization} to infer the causal dependencies. However, the lagged temporal dependence exits only when the granularity of the monitored metrics is as fine as a millisecond or second in cloud systems~\cite{aggarwal2020localization}, and setting up such a fine-grained monitoring system is quite costly.} By contrast, BALANCE can still be useful when the graph structure is not available and may in turn assist in graph construction and the subsequent localization step. 

\subsection{Attribution methods}
\label{ssec:attribution}
Since we use attribution methods to solve the problem of RCA, we review some state-of-the-art (SOTA) attribution methods in this section. The goal of attribution methods is to understand and explain why a model makes a certain prediction, thus assisting in winning user trust and further providing insight into how to enhance the performance of the model. Generally speaking, attribution methods can be divided into gradient-based and perturbation-based methods. Gradient-based methods compute the attribution values by leveraging the gradients of the model. The first attempt in this direction is to compute the absolute gradient of the target output of a model w.r.t. (with regard to) the input, which is also known as sensitivity analysis~\cite{simonyan2013deep}. However, sensitivity analysis is typically quite noisy and discards the information on the direction of the input change. One appealing solution is to use the element-wise multiplication of the gradient and the input (i.e., gradient$\times$input), in order to increase the sharpness of attribution maps~\cite{shrikumar2017learning}. The major drawback of the above naive uses of the gradient information on highly non-linear models is that only the information about the local behavior of the function in the neighbor of a given input is provided. To address this problem, many methods are proposed, including Layer-wise Relevance Propagation (LRP)~\cite{bach2015pixel},  Integrated Gradients~\cite{sundararajan2017axiomatic}, and DeepLIFT~\cite{shrikumar2017learning}. On the other hand, perturbation-based methods obtain the attribution of an input feature by removing or altering it, and then measuring the difference between the output before and after the perturbation is added. Feature occlusion~\cite{zeiler2014visualizing} directly removes each feature in turn and so requires $p$ model evaluations, where $p$ is the number of features. To reduce the heavy computational burden, Local Interpretable Model-agnostic Explanations (LIME)~\cite{ribeiro2016should} resorts to group-wise feature occlusion. These groups are used to fit a local Lasso regression and the resulting coefficients are regarded as attributions. Unfortunately, both feature occlusion and LIME suffer from the pitfall of the high sensitivity to the choice of hyper-parameters in the model. In other words, these methods may be influenced by the user-defined parameters in an unpredictable fashion and there is no guarantee that the explanation is unbiased and faithful. One remedy to this problem is to use SHapley Additive exPlanations (SHAP)~\cite{lundberg2017unified}, taking advantage of the classical Shapley values to assign credit to participants in a cooperative game. Shapley values are proven to be the only consistent attribution approach with several unique axioms in agreement with human intuition~\cite{sun2011axiomatic}. Nevertheless, it results in a daunting computational complexity of $\gO(2^p)$. This has sparked recent research efforts toward reducing the complexity~\cite{wang2022accelerating}. In BALANCE, we specify the model between the candidate root causes and the target KPIs to be linear. As a result, almost all aforementioned attribution methods can be unified~\cite{ancona2019explaining}, and hence, the proposed model enjoys various nice properties, as will be discussed in the subsequent sections.

\subsection{Sparse linear models for feature selection}
\label{ssec:linear_models}

As we employ a sparse linear forward model in BALANCE, we briefly review the relevant literature in this section. The most popular sparse linear model is Lasso~\cite{tibshirani1996regression}, which intends to find the coefficients that can best describe the linear relationship between the observed features and the outputs while enforcing the coefficients to be sparse via the $\ell_1$-norm penalty. Three problems stand in the way of a direct application of Lasso to RCA: First, there often exist correlations between the candidate causes but Lasso pales in tackling correlated features due to the nature of the $\ell_1$ norm~\cite{zou2005regularization}. Second, the tuning parameter in front of the $\ell_1$ norm, which balances the trade-off between data fidelity and coefficient sparsity is unknown in practice and is typically chosen by cross-validated grid search. As a result, the lasso algorithm has to be run for every candidate value of the tuning parameter for every partition of the dataset, leading to a heavy computational burden. Third, missing values are the rule rather than the exception, but Lasso cannot deal with them directly. As a remedy to the first problem, E-Net (Elastic net) is proposed by using the combination of the $\ell_1$ and the $\ell_2$ norm on the coefficients, which introduces the grouping effect to the correlated coefficients~\cite{zou2005regularization}. This merit comes along with an extra tuning parameter and considerable computational overhead. On the other hand, the other two issues can be addressed by Bayesian models, such as ARD (Automatic Relevance Determination)~\cite{tipping2001sparse}. The tuning parameter is assumed to be a random variable and its posterior distribution can be inferred from the data via expectation maximization. Unfortunately, correlations between features are not considered in this model. A handful of works are further proposed to alleviate this deficiency by borrowing the strength from the correlated shrinkage priors, such as the group inverse-Gamma Gamma prior~\cite{boss2021group} and the correlated spike-and-slab prior~\cite{rovckova2014negotiating}. However, such methods require some prior information that is unavailable in practice, such as the grouping information~\cite{boss2021group} or the prior distribution for zero coefficients~\cite{rovckova2014negotiating}. In this work, we propose BMFS to overcome the abovementioned issues, which will be discussed in detail in $\S$\ref{ssec:BMFS}.

\section{Problem Formulation}
\label{sec:desiderata}

\begin{figure}[]
  \centering
  \includegraphics[width=0.97\linewidth]{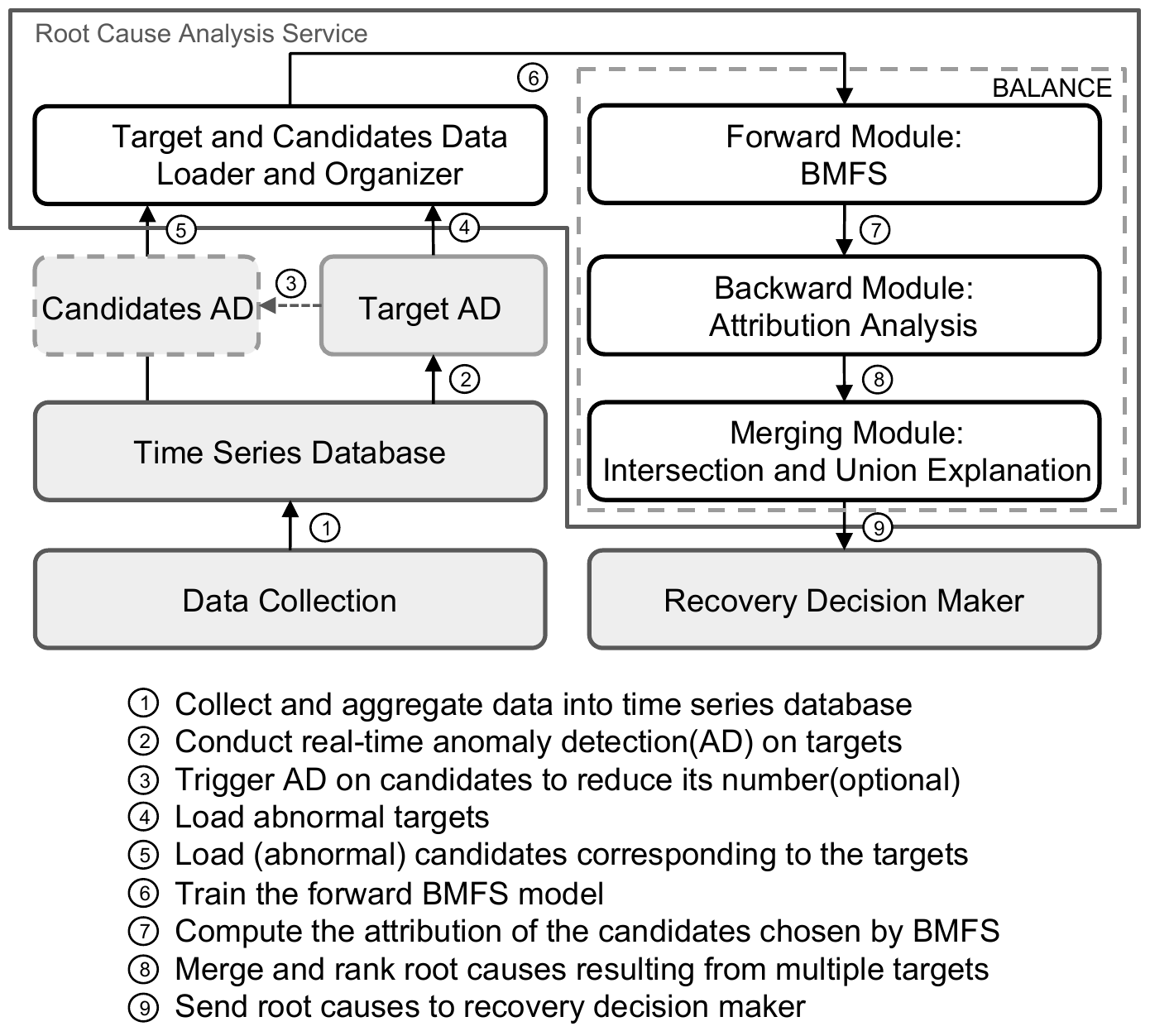}
  \vspace{-3ex}
  \caption{The overall framework of BALANCE.}
  \label{fig:larc-framework}
  \vspace{-3ex}
\end{figure}

In this section, we first introduce the overall framework of BALANCE. We then delve into in the RCA part, provide some desiderata, and discuss how such desiderata conceive the proposed method.

The overall framework of BALANCE is depicted in Figure~\ref{fig:larc-framework}. We first collect the raw data, including both target KPIs and the candidate root causes, and store them in the time series database. Concretely, let $\mX = [\vx_1, \vx_2, \cdots, \vx_p]$ denote the $p$ candidate root causes or the fundamental metrics \yh{(e.g., metrics for each SQL)} that are associated with the target KPI or the derived metric $\vy$ \yh{(e.g., tenant KPIs)}. \yh{Real-time anomaly detection is only employed to monitor the target KPI $\vy$ since it is often prohibitively resource-consuming to monitor all the fundamental metrics (i.e., the candidate root causes) $\mX$. Although not real-time, the anomaly detection on $\mX$ can be triggered once an alarm on $\vy$ is raised, since we only need to focus on the abnormal $\mX$ during RCA.} We then collect the data of the abnormal $\vy$ and $\mX$ before and during the alarm with length $n$ from the database, and input the data into the proposed RCA service, BALANCE, in order to find the root cause. The resulting estimated root cause serves as an input to the recovery decision maker, which yields the self-healing plan.

Ideally, we would like an RCA module that satisfies the following desiderata:
\begin{itemize}[leftmargin=2em,itemsep= 2pt,topsep = 2pt,partopsep=2pt]
\item[\textit{d1}.]\label{chall1} \yh{The number of observations (i.e., the length of the time series) $n$ is typically small since we only consider the short time series before and during the anomalies to explain the abnormal behavior of the targets $\vy$.}
\item[\textit{d2}.]\label{chall2} \yh{The number of candidates $p$ is varying, and possibly large in each case. Therefore, the RCA module should be sufficiently flexible to deal with the cases where $p \leq n$ and $p > n$. The latter scenario is more challenging and can be found in practice, for instance, the number of running SQLs is time-varying in the bad SQL example, and can be hundreds or even thousands, whereas $n = 61$. In this case, $p > n$.}
\item[\textit{d3}.]\label{chall3} The RCA module should output all possible root causes even if they are correlated with each other while removing all irrelevant candidates at the same time. \yh{In practice, it is infeasible to enumerate and verify all possible causalities between the candidates, since their interactions are often time-varying and bi-directional. As such, it is better to list all correlated root causes and provide the SREs with more information.} 
\item[\textit{d4}.]\label{chall4} The RCA module should be able to deal with missing data that often exist in the monitor system.
\item[\textit{d5}.]\label{chall5} The process of RCA should be efficient.
\item[\textit{d6}.]\label{chall6} The results should be interpretable and the importance of the candidates should be comparable and can be used for further ranking and decision-making.
\end{itemize}

On the other hand, attribution methods target producing explanations of the output behavior of a model by assigning a scalar attribution value $r_j$, sometimes also called ``relevance'', ``feature importance'', or ``contribution'', to each input feature $j$ of the model. Formally, given a target output $\vy$, the objective of an attribution method is to determine the contribution $\vr = [r_1, \cdots, r_p] \in \sR^p$ of each input feature $\vx_j \in \mX$ to the output $\vy$. It is therefore straightforward to exploit attribution methods for RCA. Under the framework of attribution, first, a forward model $\vy = f(\mX)$ is constructed in order to predict $\vy$ given $\mX$. Then an attribution method is used to explain the abnormal behavior of the target KPI $\vy$ by attributing it to the candidate root causes $\mX$. In this work, we advance to use a Bayesian sparse linear model as the forward model, with special attention to the correlation between the candidates. We would like to motivate the use of the forward model from the following three perspectives:
\begin{itemize}[leftmargin=1.5em,itemsep= 2pt,topsep = 2pt,partopsep=2pt]
    \item First, the proposed forward model can well capture all the above desiderata. Specifically, due to the efficiency of linear models, we fit a different linear model to the data every time the target KPIs are alarmed, successfully solving the problem of the case-varying number of candidates (i.e., \textit{d2}). \yh{By leveraging sparsity in the regression coefficients, sparse linear models can choose candidates relevant to the target in an automatic way, and can further handle the large-$p$-small-$n$ problem (i.e., \textit{d1, d2}).} After further considering the multicollinear relationship between the candidates, \textit{d3} can be satisfied. The Bayesian framework in the proposed model facilitates the processing of the missing data by inferring their distributions along with the remaining parameters (i.e., \textit{d4}). By learning the variational posterior distribution of the tuning parameters rather than estimating them via grid search, the proposed model is typically more efficient than the commonly used frequency counterpart (i.e., \textit{d5}). Finally, linear models are quite amenable to attribution methods as will be discussed below (i.e., \textit{d6}).
    \item Second, the results (i.e., the estimated root causes) yielded by an attribution method should be faithful to the underlying process the SREs are trying to understand. To this end, the methods reviewed in $\S$\ref{ssec:attribution} typically adopt a complicated forward model (i.e., a neural network) to predict $\vy$ given $\mX$, in order to maximize the predictive accuracy. On the other hand, due to the black-box nature of the forward model, another interpretable model is further applied to approximate what the forward model has learned, so as to maximize the descriptive accuracy. For instance, both LIME~\cite{ribeiro2016should} and SHAP~\cite{lundberg2017unified} employ local linear explanation models, and DeepLIFT~\cite{shrikumar2017learning} linearizes non-linear components of a neural network. Under the RCA case, it is intractable to train a neural network as the forward model due to \textit{d2}, \textit{d1}, and \textit{d5}. Moreover, our objective is not to predict $\vy$ given $\mX$, and hence, the predictive accuracy is not our main focus. Instead, we are more concerned with the descriptive error. By specifying the forward model to be linear, we minimize the descriptive error to zero. 
    \item Finally, as mentioned in $\S$\ref{ssec:attribution}, Shapley values are justified as the only possible attribution method that satisfies all axioms that are consistent with human intuition, and it has been proven in~\cite{ancona2019explaining} that almost all aforementioned attribution methods, including gradient$\times$input, integrated gradients, DeepLIFT, and feature occlusion generate exact Shapley values when applied to a linear model and a zero baseline is used. We will discuss how the proposed linear model fulfills all the axioms in $\S$\ref{ssec:back_attribution}. 
\end{itemize}

After the forward model is constructed, we then attribute the anomaly in $\vy$ to the candidates $\mX$ by finding the subset of $\mX$ that contribute the most to the abnormal behavior of $\vy$. Finally, we rank the root causes based on the attribution score and merge the results when there are multiple target KPIs. 

\section{Methodology}

In this section, we elaborate the three modules in BALANCE individually.

\subsection{Forward Module: Bayesian multicollinear Features Selection}
\label{ssec:BMFS}

\subsubsection{Bayesian Formulation}

As mentioned in $\S$\ref{sec:desiderata}, we choose a sparse linear model as the forward model, that is, 
$\vy = \mX\vbeta + \vepsilon$, where $\vy \in \sR^n$ denotes the time series of length $n$ for a single target KPI around the anomaly, $\mX \in \sR^{n\times p}$ denotes the time series of the corresponding $p$ candidate root causes, $\vbeta \in \sR^p$ is the sparse coefficient vector, and $\vepsilon\sim\gN(0, \alpha^{-1}\mI)$ represents the independent Gaussian noise with variance $\alpha^{-1}$. Equivalently, we can formulate the problem from a probabilistic perspective:
\begin{align}
\small
    p(\vy|\vbeta, \mX, \alpha) &= \gN(\mX\vbeta, \alpha^{-1}\mI) \notag\\
    &\propto \exp\left(\frac{\alpha}{2}(\vy - \mX\vbeta)^T(\vy - \mX\vbeta)\right).
\end{align}
We assume that $\alpha$ follows a non-informative Jeffrey's prior, that is, $p(\alpha) \propto 1 / \alpha$.

\begin{figure}[]
\centering
\subfigure{
\includegraphics[width=0.65\columnwidth,origin=c]{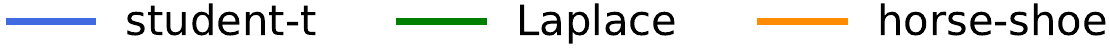}}\\
\vspace{-6pt}
\clearsubcaptcounter
\subfigure[Sparse promoting priors]{
\includegraphics[height = 0.46\columnwidth]{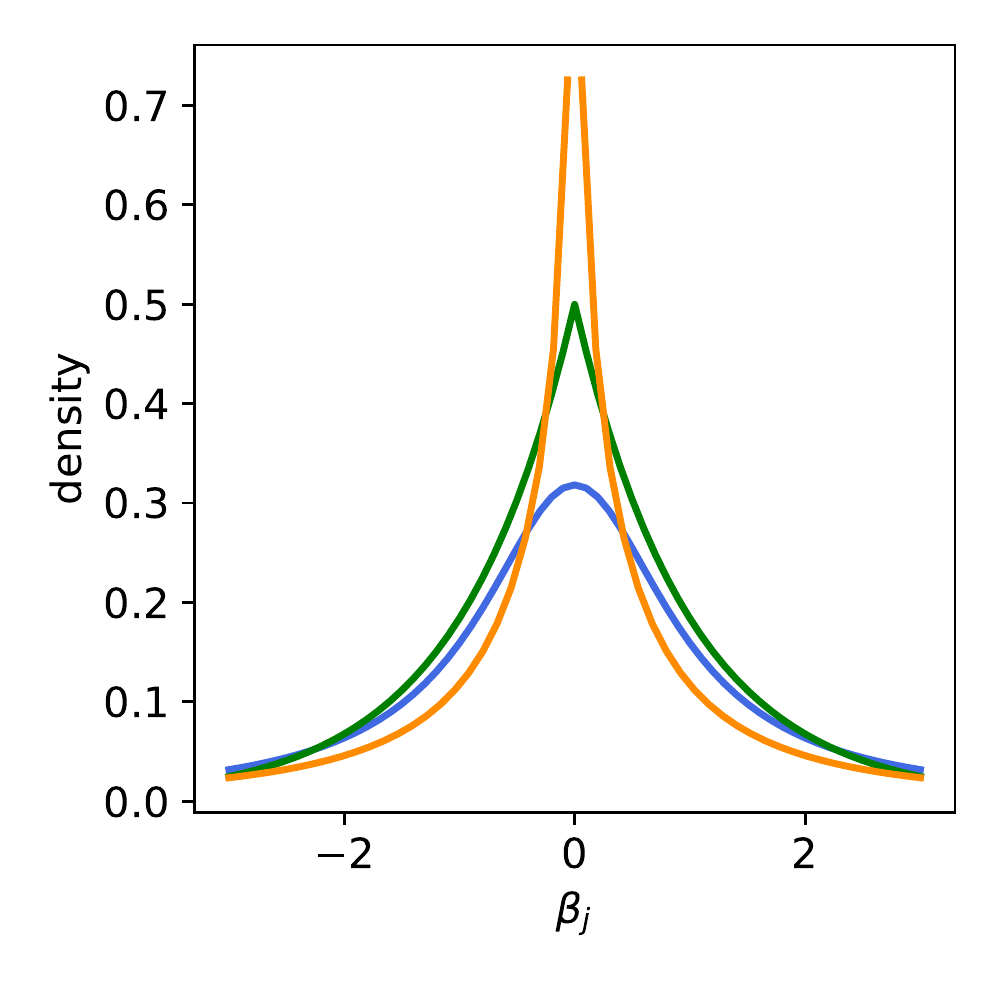}
\label{sfig:shrinkage_priors}
}
\subfigure[Shrinkage weights density]{
\includegraphics[height = 0.46\columnwidth]{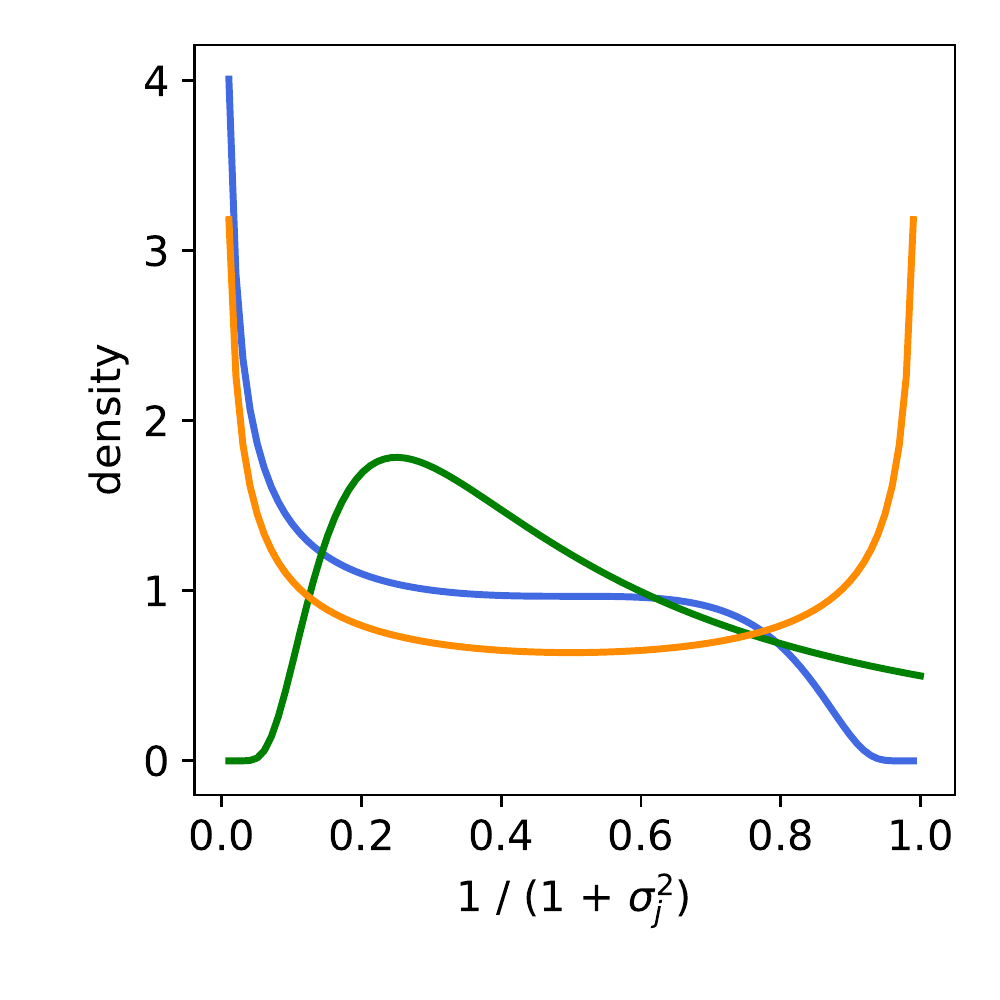}
\label{sfig:shrinkage_weights}
}
\vspace{-2ex}
\caption{The density of the commonly used sparse promoting priors, including the Student's T, Laplace, and horse-shoe prior (a), and the density of the corresponding shrinkage weights $1 / (1 + \sigma^2)$ (b).}
\label{fig:HS_prior}
\vspace{-3ex}
\end{figure}

Now let us focus on the prior distribution on $\vbeta$. Our objective is to simultaneously encourage sparsity and handle the multicollinearity among the features $\mX$. To promote sparsity, one typically resorts to the Laplace distribution (i.e., the Bayesian counterpart of the $\ell_1$-norm) or the student's T distribution. However, here we choose the horse-shoe prior~\cite{carvalho2009handling} due to its advantages over the Laplace and the student's T prior. The horse-shoe prior can be expressed as a scale mixture of Gaussians:
\begin{align}
\small
    p(\beta_j | \sigma_j, g) &= \gN(\vzero, g\sigma_j^2), \\
    p(\sigma_j) &= C^+(0, 1),
\end{align}
where $C^+(0, 1)$ is a standard half-Cauchy distribution on positive reals $\sR^+$, and $g$ and $\sigma_j$ are respectively the global and local shrinkage parameters. As $g$ becomes small, the global shrinkage parameter $g$ shrinks all the coefficients $\beta_j$ in $\vbeta$ to zero. On the other hand, the heavy-tailed half-Cauchy priors for the local shrinkage parameters $\sigma_j$ allow some $\beta_j$ to escape from the shrinkage. Therefore, the resulting $\vbeta$ would be sparse. In comparison with other commonly-used sparse promoting priors, such as the Student's T and Laplace distribution, we can tell from Figure~\ref{sfig:shrinkage_priors} that horse-shoe prior has a larger density for 0 and for very large $\beta_j$. In other words, it can better separate the zero and non-zero elements in $\vbeta$. Viewed another way, given that $\sigma_j$ follows a half-Cauchy prior on $\sigma_j$, we can obtain that~\cite{carvalho2009handling}:
\begin{align}
\small
    \frac{1}{1 + \sigma_j^2} \sim \text{Be}\Big(\frac{1}{2}, \frac{1}{2}\Big), \label{eq:shrink_weight}
\end{align}
where $\text{Be}(0.5, 0.5)$ denotes a beta distribution with shape parameters $0.5$, and we refer to $1 / (1 + \sigma^2)$ as the shrinkage weight. As shown in Figure~\ref{sfig:shrinkage_weights}, the density function of the shrinkage weight for the horse-shoe prior (i.e., the orange line) has a U-shape (i.e., horse-shoe shape); it reaches the lowest value at $0.5$ but is unbounded at $0$ and $1$, indicating that this prior prefers $\sigma_j^2$ to be either very small or very large. However, the student's T and Laplace prior (see the blue and green lines) do not have this nice property. As a result, the horse-shoe prior is more robust when dealing with unknown sparsity and large non-zero elements.

On the other hand, to tackle the multicollinearity, we typically resort to the g-prior~\cite{zellner1986assessing}:
\begin{align}
\small
    p(\vbeta| g, \sigma^2, \mX) = \gN\left(\vzero, g\sigma^2(\mX^T\mX)^{-1}\right),
\end{align}
where $g$ and $\sigma^2$ are scalars that determine the overall variance of $\vbeta$, and the inter-dependencies among different features $\mX$ are characterized empirically by $\mX^T\mX$. The resulting posterior distribution on $\vbeta$ presents a larger correlation among elements in $\vbeta$ than the posterior given by other prior distributions, such as the Laplace (i.e., $\ell_1$-norm) and the Gaussian with a diagonal covariance (i.e., $\ell_2$-norm). To bring together the best of both worlds, we propose a novel prior distribution that seamlessly integrates the horse-shoe prior and the g-prior:
\begin{align}
\small
    p(\vbeta|g, \vsigma, X) &= \gN\left(\vzero, g\diag(\vsigma)\mX^T\mX\diag(\vsigma)\right), \\
    p(\sigma_j) &= C^+(0, 1),
\end{align}
where $\vsigma$ is a $p$-dimensional vector. We name the above prior correlated horse-shoe prior. To facilitate the derivation of the variational inference algorithm, we further reparameterize the above prior using the canonical exponential family form~\cite{neville2014mean} as:
\begin{align}
\small
    p(\vbeta|\gamma, \vlambda, \mX) =\ & \gN\left(\vzero, \Big(\gamma\diag\big(\vlambda^{\frac{1}{2}}\big)\mX^T\mX\diag\big(\vlambda^{\frac{1}{2}}\big)\Big)^{-1}\right) \notag \\
    \propto\ & \exp\bigg(\frac{p}{2}\log\gamma + \frac{1}{2}\sum_{j=1}^p\log\lambda_j \notag\\
    &\qquad -  \frac{\gamma}{2} \vlambda^{\frac{T}{2}} \big((\mX^T \mX) \circ (\vbeta\vbeta^T)\big)\vlambda^{\frac{1}{2}} \bigg), \\
    p(\lambda_j) =\ & \frac{1}{\pi} \lambda_j^{-\frac{1}{2}}(\lambda_j + 1)^{-1}, \quad \forall \lambda_j > 0, 
\end{align}
where $\gamma = 1/g$, $\lambda_j = 1/\sigma_j^2$, $\vlambda ^{1/2}$ denotes element-wise square root of the entries in vector $\vlambda$, $\vlambda ^{T/2}$ is the transpose of $\vlambda ^{1/2}$, and $p(\lambda_j)$ can be regarded as a Beta prime distribution or a compound Gamma distribution. Since we do not have any prior knowledge of the global shrinkage parameter $\gamma$, we employ the non-informative Jeffrey's prior, namely, $p(\gamma) \propto 1 / \gamma$.

\begin{figure}[]
\centering
\includegraphics[width=0.5\columnwidth]{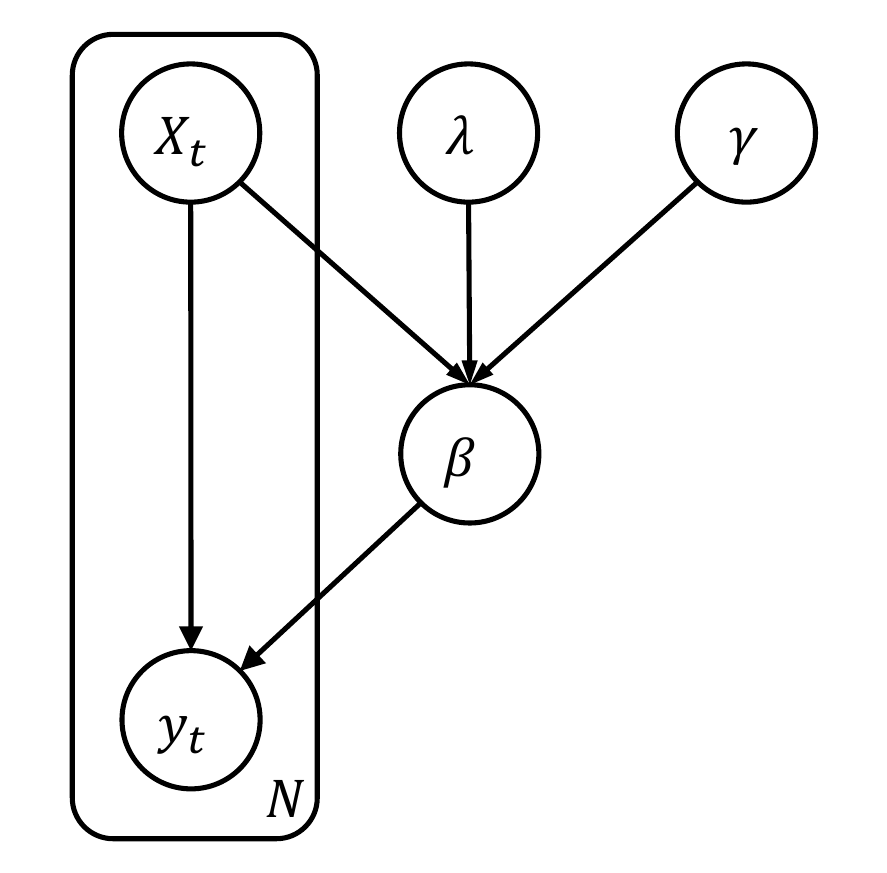}
\vspace{-2ex}
\caption{Graph representation of BMFS. The box labeled $N$ denotes that the number of repetitions for the subgraph inside is $N$.}
\label{fig:graph_BMFS}
\vspace{-3ex}
\end{figure}

Altogether, the overall Bayesian model can be factorized as:
\begin{align}
\small
    p(\vy, \vbeta, \vlambda, \alpha, \gamma | \mX) = p(\vy|\mX, \vbeta, \alpha) p(\vbeta|\gamma, \vlambda, \mX)p(\vlambda) p(\gamma)p(\alpha). 
\end{align}
The corresponding graph representation can be found in Figure~\ref{fig:graph_BMFS}. 

\subsubsection{Variational Inference}

Our overarching goal is to infer the posterior distribution $p(\vbeta, \vlambda, \alpha, \gamma |\vy, \mX)$. However, it is intractable to obtain the close-form expression of this posterior distribution, and thus, we follow the framework of variational inference and find a tractable variational distribution $q(\vbeta, \vlambda, \alpha, \gamma)$ that is closest in Kullback-Leibler (KL) divergence to the exact posterior. For simplicity, we apply the mean-field approximation and factorize the variational distribution as:
\begin{align} \label{eq:mean_field}
\small
    q(\vbeta, \vlambda, \alpha, \gamma) = q(\vbeta)\prod_{j=1}^p q(\lambda_j)q(\alpha)q(\gamma).
\end{align}
\yh{One advantage of the mean-field approximation is that the functional form of each factor can be specified by equating the functional derivatives of the KL divergence w.r.t. the factor to zero~\cite{bishop2006pattern}. After obtaining the functional form of the variational distribution, we then update the parameters of these distributions recursively via natural gradient descent~\cite{lin2019fast, yu2019variational}.} It is worthwhile to emphasize that natural gradients typically result in simpler expressions (i.e., less computation in each iteration) and faster convergence (i.e., fewer iteration numbers) than standard gradients~\cite{lin2019fast, yu2019variational, yu2020fast}. Thus, it is favored when we are concerned with the efficiency of the algorithm. \yh{In addition, although we assume the variational distributions are independent in Eq-\ref{eq:mean_field}, their parameters are dependent on each other in a straightforward way when optimizing them to minimize the KL divergence, as demonstrated in the following update rules of these parameters. In other words, we still capture the interactions between the variables $\vbeta$, $\vlambda$, $\alpha$, and $\gamma$ to some extent.} Now let us delve into the derivations for each variational distribution. 

For $q(\vbeta)$ we can specify it to be a Gaussian distribution when $\beta_j\in\sR$ or a log-normal distribution when $\vbeta$ are constrained to be non-negative. In the first scenario, $q(\vbeta)$ that minimizes the KL divergence can be updated as:
\begin{align} \label{eq:q_beta}
\small
    q(\vbeta; \vh_\beta, \mJ_\beta) \propto \exp\left(-\frac{1}{2} \vbeta^T \mJ_\beta \vbeta + \vh_\beta^T \vbeta\right),
\end{align}
where $\mJ_\beta$ and $\vh_\beta$ respectively represent the precision matrix (i.e., the inverse covariance) and the potential vector of $\beta$. Gaussian distributions parameterized by the precision matrix and the potential vector are called the canonical form and such form is amenable to concise update rules as shown below~\cite{lin2019fast, yu2019variational}. $\vh_\beta$ and $\mJ_\beta$ are known as the canonical or natural parameters, and the corresponding mean parameters can then be computed as $\langle\vbeta\rangle = \mJ_\beta^{-1}\vh_\beta$ and $\Cov[\vbeta] = \mJ_\beta^{-1}$. \yh{Specifically, update rules for $\mJ_\beta$ and $\vh_\beta$ can be derived as:
\begin{align}
\small
    \mJ_\beta^{\{\kappa\}} =\ & (1-\rho) \mJ_\beta^{\{\kappa - 1\}} + \rho \Big(\langle \alpha\rangle \mX^T\mX \notag \\
    &\qquad + \langle\gamma\rangle
    \diag\big(\langle\vlambda^{\frac{1}{2}}\rangle\big)\mX^T\mX\diag\big(\langle\vlambda^{\frac{1}{2}}\rangle\big)\Big), \\
    \vh_\beta^{\{\kappa\}} =\ & (1-\rho)\vh_\beta^{\{\kappa - 1\}} + \rho\langle\alpha\rangle\mX^T\vy, \label{eq:update_h_beta}
\end{align}
where $\langle\cdot\rangle$ denotes the expectation w.r.t. the corresponding variational distribution and $\rho$ denotes the step size determined using the line search method (e.g., the Armijo rule)~\cite{yu2019variational}. To provide more intuition for the above derivations, we take the update rule for $\vh_\beta$ as an example. In Eq-\ref{eq:update_h_beta}, the natural gradient for $\vh_\beta$ is $\langle\alpha\rangle\mX^T\vy - \vh_\beta$. As we update $\vh_\beta$ in the direction of the natural gradient with a step size $\rho$, we can obtain Eq-\ref{eq:update_h_beta}. The update rule for $\mJ_\beta$ is derived in the same fashion, and likewise for the update rules in the sequel.}

On the other hand, when we restrict $\vbeta$ to be non-negative and use the log-normal distribution as the variational distribution, we further factorize $q(\vbeta)$ as $\prod_j q(\beta_j)$ and $q(\beta_j)$ can be written as:
\begin{align} \label{eq:q_beta_pos}
\small
    q(\beta_j; h_{\beta_j}, \zeta_{\beta_j}) &\propto \frac{1}{\beta_j}\sqrt{\zeta_{\beta_j}}\exp\left(-\frac{1}{2}\zeta_{\beta_j} \beta_j^2 + h_{\beta_j}\beta_j\right), 
\end{align}
where in each iteration $\kappa$ the natural parameters $\vh_\beta$ and $\vzeta_\beta$ can be updated as:
\begin{align}
\small
    \vh_{\beta}^{\{\kappa\}} =\ & (1-\rho) \vh_{\beta}^{\{\kappa - 1\}} + \rho \Big(- \vc_1 \circ \big(1 - 2\langle\log\vbeta\rangle\big) \notag\\ 
    &\qquad +\vc_2 \circ \big(1 - \langle\log\vbeta\rangle\big) + 1\Big),  \\
    \vzeta_{\beta}^{\{\kappa\}} =\ & (1-\rho) \vzeta_{\beta}^{\{\kappa - 1\}} + \rho (2 \vc_1 - \vc_2 ),  \\
    \vc_1 =\ & \diag\Big(\langle \alpha\rangle \mX^T\mX \notag\\
    &\qquad + \langle\gamma\rangle \diag\big(\langle\vlambda^{\frac{1}{2}}\rangle\big)\mX^T\mX\diag\big(\langle\vlambda^{\frac{1}{2}}\rangle\big) \Big)\circ \langle \vbeta ^ 2\rangle, \\
    \vc_2 =\ & \langle\alpha\rangle\mX^T\vy - \offdiag\Big(\langle \alpha\rangle \mX^T\mX \notag\\
    &\qquad + \langle\gamma\rangle \diag\big(\langle\vlambda^{\frac{1}{2}}\rangle\big)\mX^T\mX\diag\big(\langle\vlambda^{\frac{1}{2}}\rangle\big) \Big)\circ\langle\vbeta\rangle,
\end{align}
where $\circ$ is the Hadamard (or elementwise) product, $\vbeta ^ 2$ denotes elementwise square of $\vbeta$, and $\offdiag(\cdot)$ denotes the off-diagonal part of a matrix by replacing the diagonal with a zero vector. Afterwards, we can compute the mean parameters as $\langle\log\vbeta\rangle = \vh_\beta \oslash \vzeta_\beta$, $\Var[\log\vbeta] = 1 \oslash \vzeta_\beta$, $\langle\vbeta\rangle = \exp(\langle\log\vbeta\rangle + \Var[\log\vbeta] / 2)$, and $\langle\vbeta ^ 2\rangle = \exp(2\langle\log\vbeta\rangle + 2\Var[\log\vbeta])$, where $\oslash$ denotes elementwise division.

For $q(\lambda_j)$, we follow~\cite{neville2014mean} and specify its functional form to be:
\begin{align} \label{eq:q_lambda}
\small
    q(\lambda_j; d_j) = \frac{1}{E_1(d_j)}(\lambda_j + 1)^{-1}\exp\Big(-d_j\big(\lambda_j + 1\big)\Big),
\end{align}
where $E_1(x) = \int_x^\infty exp(-t)/t dt$ represents the exponential integral function. It follows that the update rule of $
\vd$ is:
\begin{align}
\small
    \vd^{\{\kappa\}} =\ & (1-\rho) \vd^{\{\kappa - 1\}} \notag\\
                    &+ \rho \langle\gamma\rangle\Big(\offdiag\big(\mX^T\mX\circ\langle \vbeta\vbeta^T\rangle\big)\langle\vlambda^{\frac{1}{2}}\rangle \circ \vc_3 \notag\\
                    &\qquad\quad + \frac{1}{2}\diag\big(\mX^T\mX\circ\langle \vbeta\vbeta^T\rangle\big)  \Big), \\
    \vc_3 =\ & \Big(\langle\vlambda^{\frac{1}{2}}\rangle\circ \vd - \Gamma(1.5)\circ \vd^{\frac{1}{2}}\Big) \oslash \big(\langle\vlambda\rangle\circ\vd - 1\big). 
\end{align}

The mean parameters $\langle\vlambda\rangle$ and $\langle\vlambda^{\frac{1}{2}}\rangle$ can be calculated as:
\begin{align}
\small
    \langle\vlambda\rangle &= \Gamma(-1, \vd) \oslash \Gamma(0, \vd), \label{eq:E_lambda}\\
    \langle\vlambda^{\frac{1}{2}}\rangle &= \Gamma(1.5) \Gamma(-0.5, \vd) \oslash \Gamma(0, \vd),\label{eq:E_lambda_0.5}
\end{align}
where $\Gamma$ represents the gamma function when there is only one input and the upper incomplete gamma function\footnote{$\Gamma(c, d) = U(1-c, 1-c, d) / \exp(d)$, where $U$ represents Tricomi's confluent hypergeometric function.} when there are two inputs, and the step size $\rho$ is again determined by the Armijo rule. 

For $q(\alpha)$ and $q(\gamma)$, we specify them to be the gamma distributions, and the corresponding update rules are:
\begin{align}
\small
    q(\alpha; a_\alpha, b_\alpha) &\propto \alpha^{a_\alpha - 1}\exp(- b_\alpha \alpha), \label{eq:q_alpha} \\
    q(\gamma; a_\gamma, b_\gamma) &\propto \gamma^{a_\gamma - 1}\exp(- b_\gamma \alpha), \label{eq:q_gamma}
\end{align}
where
\begin{align}
\small
    a_\alpha^{\{\kappa\}} =\ & (1-\rho)a_\alpha^{\{\kappa-1\}} + \frac{\rho n}{2},\\
    b_\alpha^{\{\kappa\}} =\ & (1-\rho)b_\alpha^{\{\kappa-1\}} \notag\\
    &+\frac{\rho}{2}\big(\vy^T\vy + \tr(\langle\vbeta \vbeta^T\rangle \mX^T\mX) - 2 \vy^T\mX\langle\vbeta\rangle\big), \\
    a_\gamma^{\{\kappa\}} =\ & (1-\rho)a_\gamma^{\{\kappa-1\}} +  \frac{\rho p}{2}, \\
    b_\gamma^{\{\kappa\}} =\ & (1-\rho)b_\gamma^{\{\kappa-1\}} +  \frac{\rho}{2}\Big(\langle\vlambda\rangle^T \diag\big(\mX^T\mX\circ\langle \vbeta\vbeta^T\rangle\big) \notag\\ 
            &\qquad+\langle\vlambda^{\frac{1}{2}}\rangle^T\offdiag\big(\mX^T\mX\circ\langle \vbeta\vbeta^T\rangle\big)\langle\vlambda^{\frac{1}{2}}\rangle\Big). 
\end{align}

In practice, entries in $\vy$ and $\mX$ are often missing. In this case, we generate point estimates for $X_{i,\vj}$ and Bayesian estimates for $y_i$ by minimizing the aforementioned KL divergence, that is,
\begin{align}
\small
    \hat X_{i,\vj} &= \langle\vbeta_{\vj}\vbeta_{\vj}^T \rangle^{-1}\langle \vbeta_{\vj}\rangle y_i, \label{eq:update_X}\\
    q(y_i) &= \gN\big(\mX_{i,:}\langle\vbeta\rangle,\langle\alpha\rangle^{-1}\big), \label{eq:update_y}
\end{align}
where the vector $\vj$ denotes the indices of the missing values in row $i$ in $\mX$, and $\mX_{i,:}$ denotes the $i$-th row of $\mX$.

\subsubsection{Soft Thresholding}
\label{sssec:soft_threshold}

\begin{figure}[t]
\centering
\vspace{-2ex}
\includegraphics[width=0.85\columnwidth]{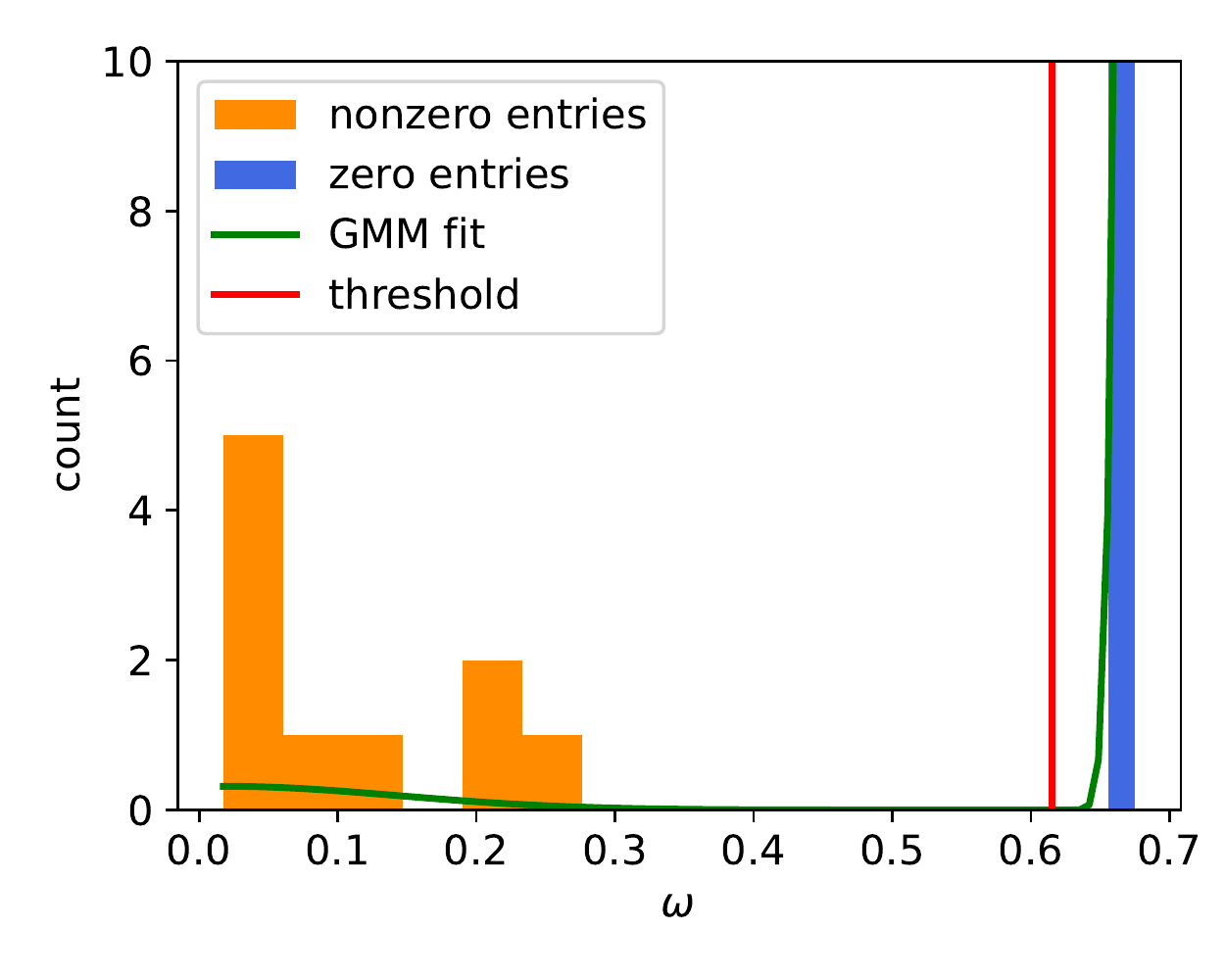}
\vspace{-3.5ex}
\caption{The empirical distribution of $\omega$ for a synthetic dataset: the orange and blue bars denote the histogram of $\omega$ w.r.t. nonzero and zero entries in $\langle\vbeta\rangle$ respectively, the green line denotes the density after fitting a GMM to the empirical distribution, and the red line denotes the chosen threshold $\hat\omega$.}
\label{fig:emp_omega_dist}
\vspace{-2.5ex}
\end{figure}

Recall that the half-Cauchy prior on $\sigma_j$ leads to the U-shaped prior of the shrinkage weight $1 / (1 + \sigma_j^2)$ (cf. Eq-\ref{eq:shrink_weight}). Since $\lambda_j = 1 / \sigma_j^2$, the shrinkage weight can be equivalently expressed as $\lambda_j / (1 + \lambda_j)$. This U-shaped prior constraints $\lambda_j$ to be either very small or very large and can separate the zero and non-zero entries in $\vbeta$ in an automatic fashion. Owing to this prior, we observe that the density of the empirical distribution $\hat p(\omega)$ on $\omega = \langle\lambda_j\rangle / (\langle\lambda_j\rangle + 1)$ for all $j$ also follows a U-shape approximately, where the expectation $\langle\lambda_j\rangle$ is taken over the variational posterior distribution $q(\lambda_j)$, as shown in Figure~\ref{fig:emp_omega_dist}. As expected, the value of $\omega$ for most true non-zero entries is close to zero and its density decreases with $\omega$, whereas the value of $\omega$ for most true zero entries is far away from zero.

To distinguish the non-zero entries from the zero ones, we choose the threshold to be $\hat\omega = \argmin \hat p(\omega)$, which is the valley of the U-shape density and set $\langle\beta_j\rangle = 0$ if $\langle\lambda_j\rangle / (\langle\lambda_j\rangle + 1) > \hat\omega$. Specifically, we fit a two-component Gaussian mixture model (GMM) to $\hat p(\omega)$: the means of the two Gaussians are fixed to the smallest and largest value of $\omega$ across all $j$ respectively, while the variances and the weights are learned from the data. We then find the $\hat\omega$ corresponding to the minimum of the density of the GMM.

We summarize the entire procedure of BMFS in Algorithm~\ref{alg:BMFS}.

\subsection{Backward Module: Attribution Analysis}
\label{ssec:back_attribution}

Since our forward model is a linear model, one tempting choice for attribution score is the regression coefficient, that is,
\begin{align}
 \small
    r_j =| \langle\beta_j\rangle |, \label{eq:sensitivity}
\end{align}
where $r_j$ denotes the attribution score and $|\cdot|$ denotes the absolute value. In fact, for linear models, the regression coefficient $\langle\beta_j\rangle$ equals the gradient of $\vy$ w.r.t. $\vx_j$, thus it describes the sensitivity of $\vy$ to $\vx_j$. The value $\langle\beta_j\rangle$ quantizes the impact of a small change in $\vx_j$ to $\vy$. Note that we have a distribution for $\beta_j$ instead of a point estimate. Following the framework of XAI for Bayesian models~\cite{bykov2021explaining}, we use the mean of $\beta_j$ here to compute the attribution score in this section. Note that other values can also be used, such as the quantiles, the modes, and the intersection or union of the modes if there is more than one mode. Since $q(\beta_j)$ follows a Gaussian distribution, the mean is the proper choice.

\begin{algorithm}[t]
\caption{Bayesian multicollinear Feature Selection (BMFS)}
\label{alg:BMFS}
\small
\begin{algorithmic}[1]
\Require an alarmed target KPI $\vy$ and candidate root causes $\mX$ associated with $\vy$;
\Ensure regression coefficients $\langle\vbeta\rangle$;
\State Initialize the parameters for all variatonal distributions;
\Repeat
    \If {$\vbeta$ can only take positive values}
        \State update $q(\vbeta)$ following Eq-\ref{eq:q_beta_pos};
    \Else
        \State update $q(\vbeta)$ following Eq-\ref{eq:q_beta};
    \EndIf
    \State compute $\langle\vbeta\rangle$ and $\Cov[\vbeta]$ given $q(\vbeta)$;
    \State update $q(\vlambda)$ by Eq-\ref{eq:q_lambda} and compute $\langle\vlambda\rangle$ by Eq-\ref{eq:E_lambda}, $\langle\vlambda^{\frac{1}{2}}\rangle$ by Eq-\ref{eq:E_lambda_0.5};
    \State update $q(\alpha)$ by Eq-\ref{eq:q_alpha}, $q(\gamma)$ by Eq-\ref{eq:q_gamma}, compute their expectations;
    \If {there exists missing values in $\mX$ and $\vy$}
        \State impute the missing values following Eq-\ref{eq:update_X} and Eq-\ref{eq:update_y};
    \EndIf
\Until {convergence;}
\State perform soft thresholding as introduced in $\S$\ref{sssec:soft_threshold};
\end{algorithmic}
\end{algorithm}

Unfortunately, high sensitivity does not indicate a high contribution to the anomaly in the target KPI. For example, suppose that the target KPI is a summary or aggregation of the candidate root causes and that $\langle\beta_j\rangle$ is non-zero and relatively large, but $\vx_j$ is small. We typically omit $\vx_j$ because its scale is small and cannot contribute much to the overall target. To this end, we resort to the gradient$\times$input approach, in which the attribution score can be computed as:
\begin{align}
\small
    r_j = |\langle\beta_j\rangle x_j|.
\end{align}
This value is known as salience in the literature of attribution~\cite{shrikumar2017learning}. Now the candidate root causes will be chosen only when both $\langle\beta_j\rangle$ and $x_j$ itself are relatively large. 

However, our ultimate objective is to attribute the anomalies in the target KPIs to the candidate root causes, and hence, our focus is not on the absolute value of $x_j$. Instead, we intend to explain the changes during the anomaly in $\vy$ by the changes in $\vx_j$. In other words, we are interested in the marginal effect of a candidate, and we are looking for how the target would change after replacing the abnormal part in the candidates with the normal part. Recall that in our case, the anomaly or alarm time is given. Hence, we choose a baseline as the mean or the median of the normal part of the time series and compute the difference $\Delta x_j$ between the anomaly and the baseline. The resulting attribution score is defined as:
\begin{align}
\small
    r_j = |\langle\beta_j\rangle \Delta x_j|.
\end{align}
To further guarantee the attribution score to be invariant to the scale of $\vy$, we finally calculate the attribution score as:
\begin{align}
\small
    r_j = \bigg|\frac{\langle\beta_j\rangle \Delta x_j}{\Delta y}\bigg|. \label{eq:relative_attribution}
\end{align}

It is worthwhile to emphasize that owing to the use of the proposed forward model. The above attribution score satisfies all the desirable axioms of the Shapley values, including completeness, null player, symmetry, linearity, and continuity~\cite{ancona2019explaining}. Here we briefly go through these axioms in the case of RCA and discuss how the proposed model copes with them. 
\begin{itemize}[leftmargin = 2em,itemsep= 2pt,topsep = 2pt,partopsep=2pt]
    \item Completeness is satisfied when attributions sum up to the difference between the value of $\vy$ during normal and abnormal periods. It is obvious that Eq-\ref{eq:relative_attribution} fulfills this requirement.
    \item Null players: if the target $\vy$ does not depend on some candidates $\vx_j$, then $r_j = 0$. The proposed correlated horse-shoe prior automatically excludes the irrelevant candidates $\vx_j$ and set the corresponding $\langle\beta_j\rangle = 0$. As a result, $r_j = 0$.
    \item Symmetry: if the target $\vy$ depends on two candidates $\vx_j$ and $\vx_k$ but not on their order, then $r_j = r_k$. This axiom is satisfied by considering multicollinearity in our forward model.
    \item Linearity: this property is satisfied by linear models naturally.
    \item Continuity: attributions generated for two nearly identical inputs should be nearly identical. This axiom can also be handled by capturing multicollinearity in our forward model.
\end{itemize}
Moreover, it is apparent that the above axioms are in agreement with the desiderata of an ideal RCA mentioned in $\S$\ref{sec:desiderata}. This again bolsters our belief that attribution analysis is a natural fit for RCA. 

\begin{figure*}[]
    \centering
    \subfigure{
    \includegraphics[width=0.3\linewidth,origin=c]{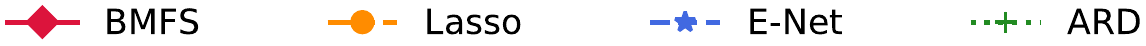}}\\
    \vspace{-1.5ex}
    \clearsubcaptcounter
    \hfill
    \subfigure[Absent]{
    \includegraphics[width = 0.3\linewidth]{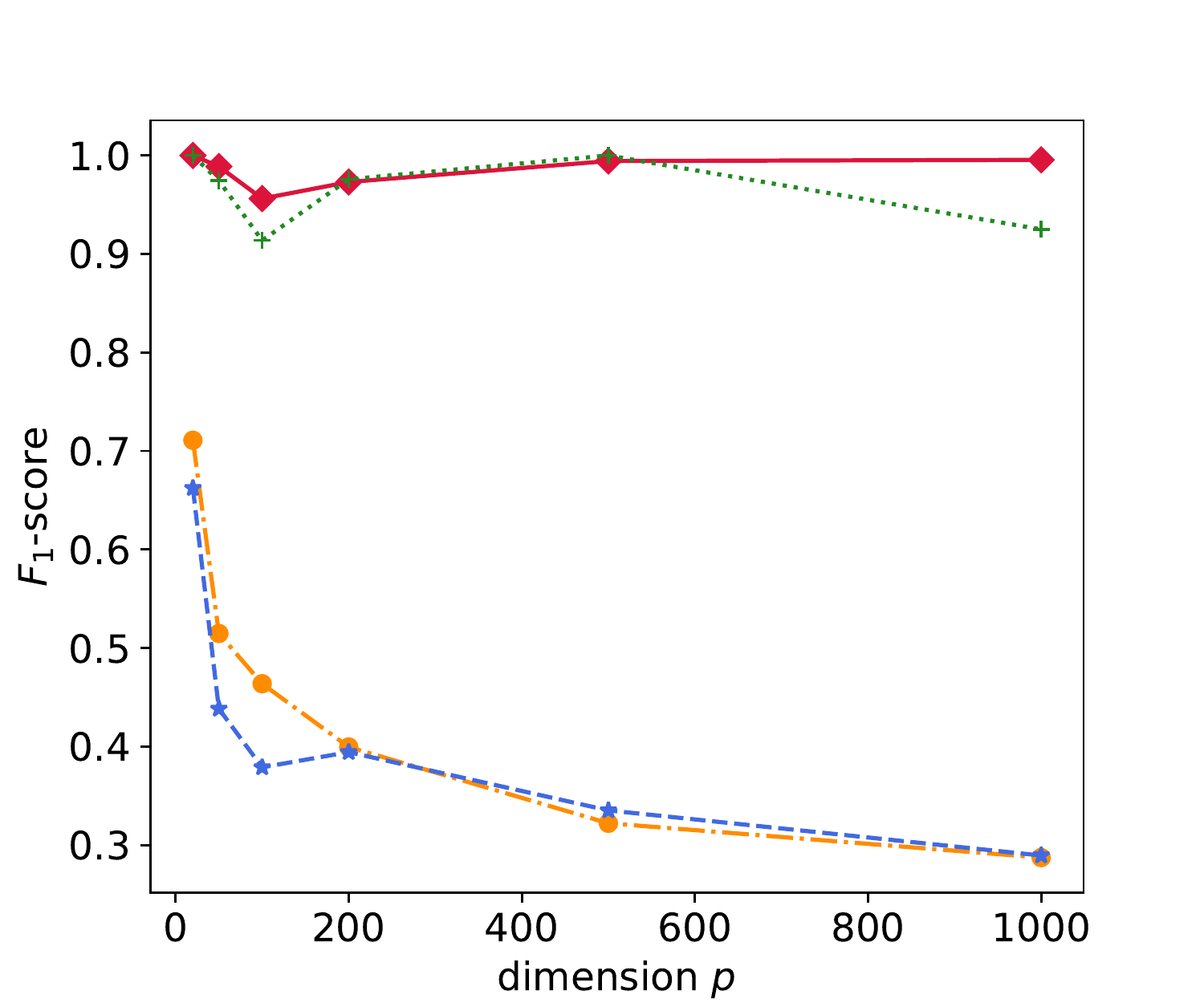}
    \label{sfig:trace_failures}
    }
    \hfill
    \subfigure[Partial]{
    \includegraphics[width = 0.3\linewidth]{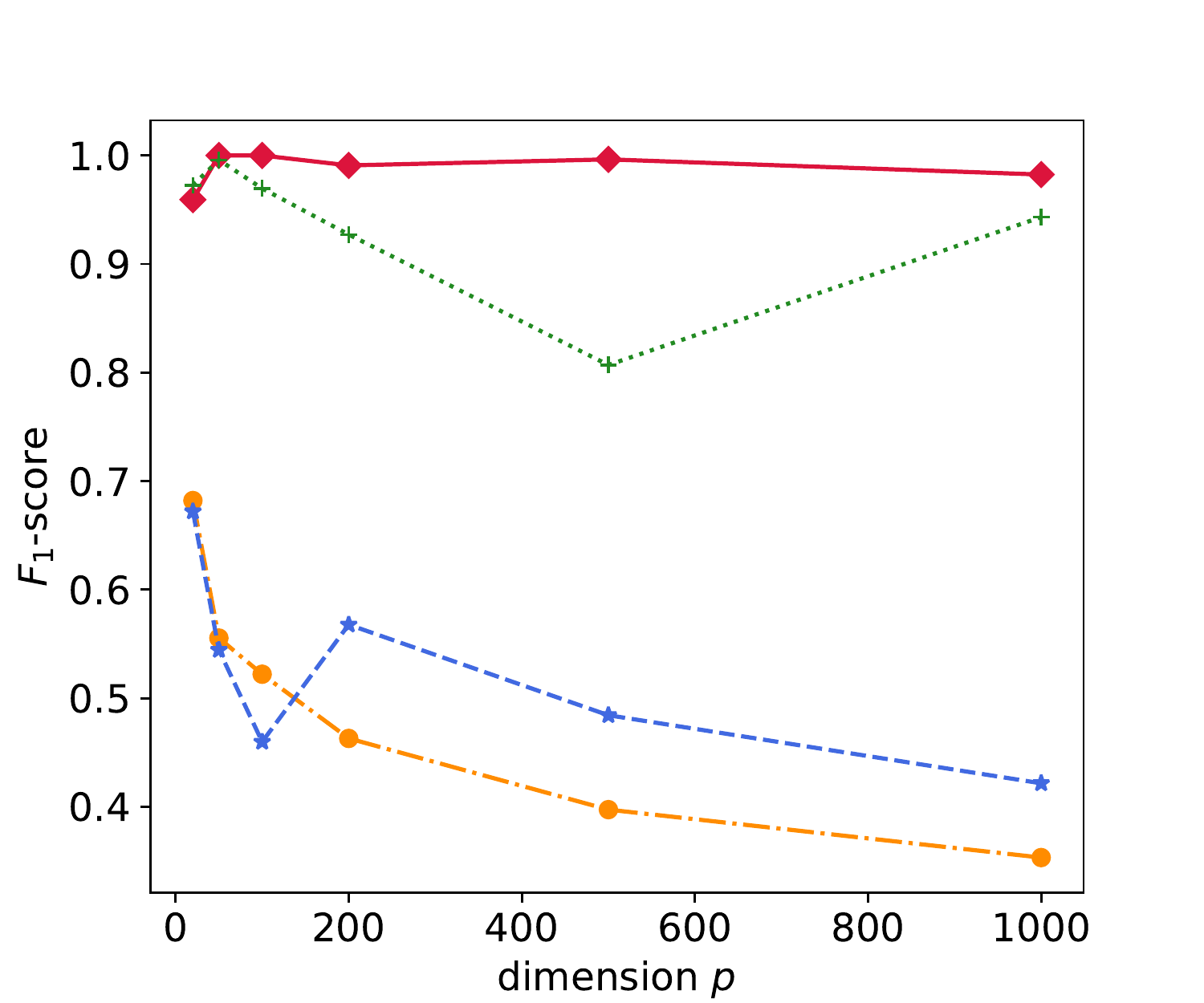}
    \label{sfig:cpu_metrics}
    }
    \hfill
    \subfigure[Perfect]{
    \includegraphics[width = 0.3\linewidth]{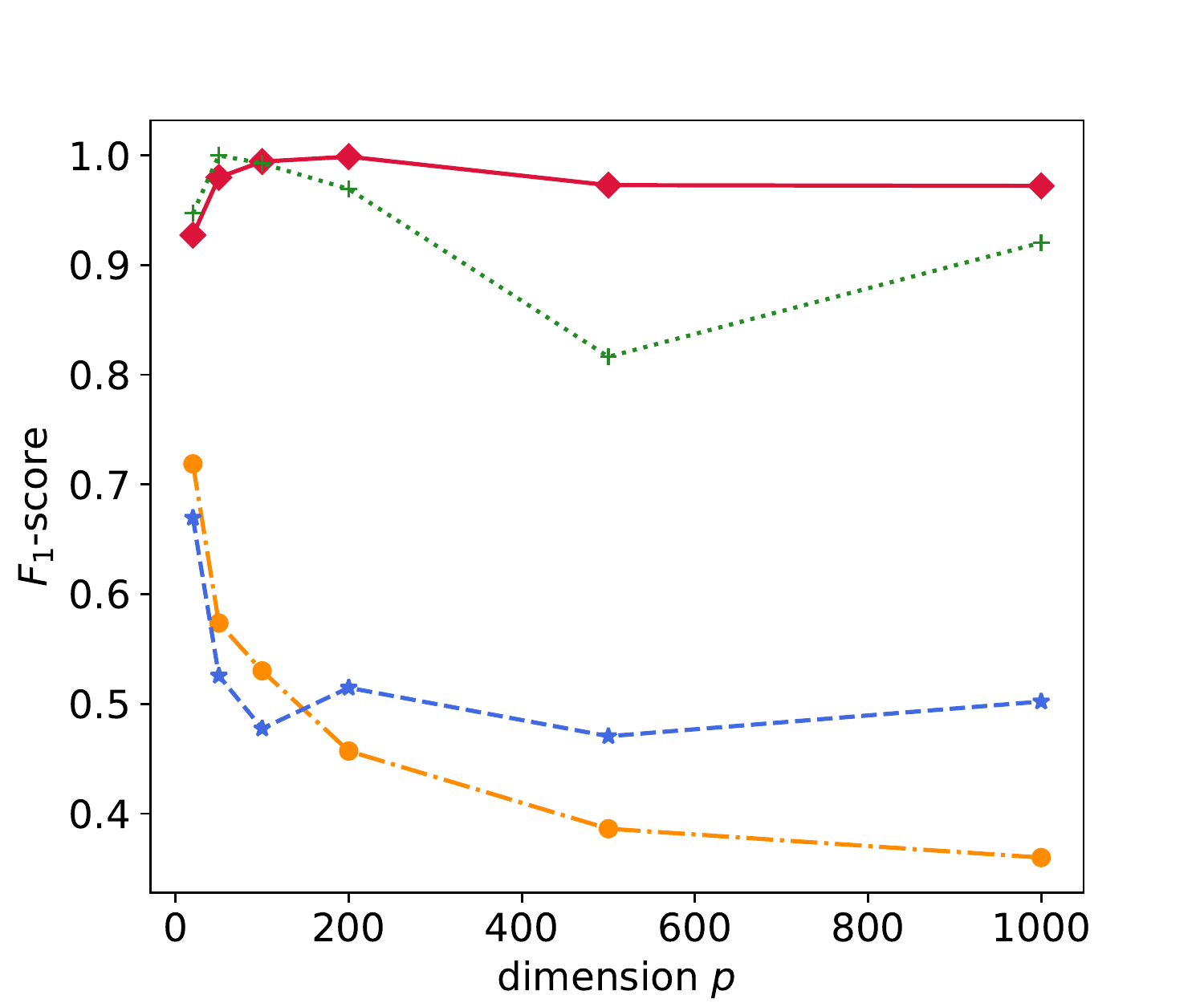}
    \label{sfig:mem_metrics}
    }
  \vspace{-3ex}
  \caption{$F_1$-score as a function of the dimension $p$ resulting from all benchmark methods in case of (a) absent of mutlicollinearity, (b) partial multicollinearity, and (c) perfect multicollinearity.}
  \label{fig:syn_mutlicolinear_dim}
  \vspace{-2.5ex}
\end{figure*}

\subsection{Merging Module: Intersection and Union Explanation}
In this subsection, we turn our attention to the case where there are multiple target KPIs. Under this situation, we first compute the attribution score $r_{jk}$ of each candidate root causes $\vx_j$ for each target KPI $\vy_k$ following the two steps in the above two subsections. We then sort the attribution scores $r_{jk}$ across $j$ in the descending order for each $k$ and refer to this step as ranking. We further retain the top $\kappa$ root causes for each target KPI $\vy_k$. Note that when $\vbeta$ is properly sparse (i.e., the number of non-zero entries in $\vbeta$ is smaller than $\kappa$), the selection step can be omitted. After ranking and selection, we merge the sets of root causes via the intersection or union of the sets. The intersection operation indicates that the root cause is chosen only when it influences all target KPIs, while the union operation detects all possible root causes that propagate abnormally to at least one target KPI.

\section{Experiments}
In this section, we first assess the performance of BALANCE on synthetic data. We then show the performance of BALANCE on three real-world applications\footnote{a. The dataset does not contain any Personal Identifiable Information (PII).
b. The dataset is desensitized and encrypted.
c. Adequate data protection was carried out during the experiment to prevent the risk of data copy leakage, and the dataset was destroyed after the experiment.
d. The dataset is only used for academic research, and it does not represent any real business situation.}, including Bad SQL localization, Container fault localization, and Fault Type Diagnosis for Exathon. The first application mainly focuses on homogeneous pairs of the target KPIs and the candidate root causes, while the latter two consider heterogeneous ones. 

\yh{\noindent\textbf{Implementation Details:}
We adopt the BALANCE framework, and for the forward module, we juxtapose our BMFS with other SOTA methods, including Lasso, E-Net, and ARD in terms of estimation accuracy and run time. The backward module is the same. For a fair comparison, we replace the correlated horse-shoe prior with the original horse-shoe prior and regard the resulting method as ARD. Thus, ARD can be regarded as an ablation study on BMFS by removing the $g$-prior part in BMFS. Note that in the original ARD~\cite{tipping2001sparse}, the Student's $T$-prior is used instead to encourage sparsity. All the compared methods have the same computation complexity of $\gO(p^2\max(n, p))$ according to our analysis. All methods are implemented in Python 3.9 except the newly added R package ``fsMTS''. All the experiments are conducted on the same Linux Server with Intel Xeon E5-2682 v4 @ 2.50GHz processors and 16GB RAM so that all the execution time could be directly compared.}

\begin{table*}[t]
\centering
\caption{Results of Synthetic Data with different levels of multicollinearity (absent, partial, and perfect) and different dimensions $p$ averaged over 100 trials. Here, $n=100$, $\alpha^{-1} = 0.01$, and proportion of nonzero coefficients $=0.5\%$.}
\vspace{-2ex}
\label{tab:syn_multicollinar_dim}
\resizebox{\textwidth}{!}{
\begin{tabular}{c|c|ccc|ccc|ccc|ccc}
\hline
 & \multirow{2}*{$p$}  & \multicolumn{3}{c|}{\bf Lasso} & \multicolumn{3}{c|}{\bf E-Net} & \multicolumn{3}{c|}{\bf ARD} & \multicolumn{3}{c}{\bf BMFS}\\
\cline{3-14}
 &  & $F_1$-score    & MSE   & Time (s)  & $F_1$-score    & MSE   & Time (s) & $F_1$-score    & MSE   & Time (s)  & $F_1$-score    & MSE   & Time (s)\\
\hline
\multirow{6}*{\rotatebox[origin=c]{90}{Absent}}
    & 20    &0.7109&6.90e-05&4.70e-01    &0.6623&7.490e-05&4.65e-01    &\textbf{0.9999}&\textbf{2.60e-05}&\textbf{2.34e-01}    &\textbf{0.9999}&\textbf{2.60e-05}&2.35e-01 \\
    & 50    &0.5148&4.10e-05&5.06e-01    &0.4383&6.60e-05&5.08e-01    &0.9743&1.50e-05&3.10e-01    &\textbf{0.9887}&\textbf{1.30e-05}&\textbf{2.72e-01} \\
    & 100   &0.4638&3.00e-05&5.84e-01    &0.3789&5.10e-05&5.88e-01    &0.9137&1.70e-05&4.47e-01    &\textbf{0.9562}&\textbf{1.20e-05}&\textbf{4.10e-01} \\
    & 200   &0.3995&1.70e-05&8.59e-01    &0.3943&3.60e-05&8.62e-01    &\textbf{0.9758}&\textbf{4.00e-06}&\textbf{3.20e-01}    &0.9730&\textbf{4.00e-06}&3.51e-01 \\
    & 500   &0.3221&9.00e-06&1.99e+00    &0.3352&2.60e-05&2.00e+00    &\textbf{0.9999}&\textbf{1.00e-06}&5.32e-01    &0.9943&\textbf{1.00e-06}&\textbf{4.94e-01} \\
    & 1000  &0.2873&5.74e-06&3.01e+00    &0.2894&1.96e-05&3.25e+00    &0.9249&1.87e-05&2.12e+00    &\textbf{0.9955}&\textbf{9.35e-07}&\textbf{2.00e+00} \\
\hline
\multirow{6}*{\rotatebox[origin=c]{90}{Partial}}
    & 20    &0.6821&4.16e-02&4.72e-01    &0.6720&2.85e-02&4.77e-01    &\textbf{0.9724}&\textbf{2.86e-02}&\textbf{2.54e-01}    &0.9593&2.87e-02&4.15e-01 \\
    & 50    &0.5554&1.90e-02&5.50e-01    &0.5445&1.32e-02&5.84e-01    &0.9958&\textbf{1.30e-02}&\textbf{2.91e-01}    &\textbf{0.9999}&\textbf{1.30e-02}&3.70e-01 \\
    & 100   &0.5221&9.36e-03&6.18e-01    &0.4598&6.48e-03&6.76e-01    &0.9696&6.32e-03&3.82e-01    &\textbf{0.9999}&\textbf{6.32e-03}&\textbf{3.36e-01} \\
    & 200   &0.4630&4.56e-03&8.52e-01    &0.5678&3.25e-03&9.52e-01    &0.9271&3.17e-03&1.06e+00    &\textbf{0.9908}&\textbf{3.16e-03}&\textbf{4.59e-01} \\
    & 500   &0.3974&1.93e-03&1.87e+00    &0.4843&1.36e-03&2.24e+00    &0.8069&1.32e-03&1.40e+00    &\textbf{0.9963}&\textbf{1.31e-03}&\textbf{6.55e-01} \\
    & 1000  &0.3532&9.19e-04&2.89e+00    &0.4216&6.45e-04&3.83e+00    &0.9431&6.22e-04&2.13e+00    &\textbf{0.9824}&\textbf{6.18e-04}&\textbf{2.09e+00} \\
\hline
\multirow{6}*{\rotatebox[origin=c]{90}{Perfect}}
    & 20    &0.7188&4.24e-02&4.74e-01    &0.6696&\textbf{2.84e-02}&4.82e-01    &\textbf{0.9474}&2.93e-02&\textbf{2.83e-01}    &0.9273&3.02e-02&4.13e-01 \\
    & 50    &0.5736&1.92e-02&5.11e-01    &0.5257&1.35e-02&5.42e-01    &\textbf{0.9999}&\textbf{1.33e-02}&\textbf{2.67e-01}    &0.9799&1.34e-02&3.61e-01 \\
    & 100   &0.5301&9.78e-03&5.83e-01    &0.4775&6.82e-03&6.39e-01    &0.9927&6.66e-03&\textbf{3.15e-01}    &\textbf{0.9944}&\textbf{6.66e-03}&3.85e-01 \\
    & 200   &0.4570&4.36e-03&7.82e-01    &0.5147&3.11e-03&9.04e-01    &0.9693&3.03e-03&5.49e-01    &\textbf{0.9988}&\textbf{3.03e-03}&\textbf{4.50e-01} \\
    & 500   &0.3862&1.52e-03&1.81e+00    &0.4705&1.07e-03&2.20e+00    &0.8164&1.03e-03&1.53e+00    &\textbf{0.9729}&\textbf{1.03e-03}&\textbf{8.11e-01} \\
    & 1000  &0.3599&7.62e-04&2.87e+00    &0.5022&5.23e-04&3.69e+00    &0.9205&5.07e-04&\textbf{2.11e+00}    &\textbf{0.9722}&\textbf{4.99e-04}&2.12e+00 \\
\hline
\end{tabular}}
\vspace{-2ex}
\end{table*}

\begin{table*}[t]
\centering
\caption{Results of Synthetic Data with different levels of noise standard deviations (noise std) averaged over 100 trials. Here, $n=100$, $p=1000$, and proportion of nonzero coefficients$=0.5\%$.} 
\vspace{-2ex}
\label{tab:syn_noise}
\resizebox{\textwidth}{!}{
\begin{tabular}{c|ccc|ccc|ccc|ccc}
\hline
 \multirow{2}*{noise std}  & \multicolumn{3}{c|}{\bf Lasso} & \multicolumn{3}{c|}{\bf E-Net} & \multicolumn{3}{c|}{\bf ARD} & \multicolumn{3}{c}{\bf BMFS}\\
\cline{2-13}
 & $F_1$-score    & MSE   & Time (s)  & $F_1$-score    & MSE   & Time (s) & $F_1$-score    & MSE   & Time (s)  & $F_1$-score    & MSE   & Time (s)\\
\hline
 0.01    &0.8466&9.07e-04&2.91e+00    &0.8820&6.150e-04&3.84e+00    &0.9250&6.16e-04&2.14e+00    &0.9789&\textbf{6.10e-04}&\textbf{2.05e+00} \\
 0.1     &0.3746&8.81e-04&2.88e+00    &0.4562&6.06e-04&3.71e+00    &0.9231&5.87e-04&\textbf{2.06e+00}    &0.9625&\textbf{5.80e-04}&2.11e+00 \\
 1.0     &0.3353&\textbf{1.07e-03}&\textbf{3.14e+00}    &0.3294&1.05e-03&4.57e+00    &0.3596&1.32e-03&3.19e+00    &0.4054&1.39e-03&3.51e+00 \\
 3.0     &0.1150&2.19e-03&3.90e+00    &0.1337&\textbf{2.09e-03}&6.19e+00    &0.0863&1.08e-02&3.38e+00    &0.0881&1.16e-02&\textbf{3.37e+00} \\
\hline
\end{tabular}}
\vspace{-2ex}
\end{table*}

\begin{table*}[t]
\centering
\caption{Results of Synthetic Data with different ratios of nonzero coefficients (nonzeros$\%$) averaged over 100 trials. Here, $n=100$, $p=1000$, and $\alpha^{-1} = 0.01$.} 
\vspace{-2ex}
\label{tab:syn_nonzeros}
\resizebox{\textwidth}{!}{
\begin{tabular}{c|ccc|ccc|ccc|ccc}
\hline
 \multirow{2}*{nonzeros$\%$}  & \multicolumn{3}{c|}{\bf Lasso} & \multicolumn{3}{c|}{\bf E-Net} & \multicolumn{3}{c|}{\bf ARD} & \multicolumn{3}{c}{\bf BMFS}\\
\cline{2-13}
 & $F_1$-score    & MSE   & Time (s)  & $F_1$-score    & MSE   & Time (s) & $F_1$-score    & MSE   & Time (s)  & $F_1$-score    & MSE   & Time (s)\\
\hline
 0.2\%    &0.4482&2.16e-04&2.95e+00    &0.4054&1.19e-04&3.53e+00    &0.9766&1.08e-04&1.97e+00    &\textbf{0.9999}&\textbf{1.07e-04}&\textbf{1.93e+00} \\
 0.5\%    &0.3927&8.20e-04&2.97e+00    &0.4838&5.87e-04&3.96e+00    &0.9296&5.67e-04&2.30e+00    &\textbf{0.9726}&\textbf{5.63e-04}&\textbf{1.94e+00} \\
 1\%   &0.3810&2.19e-03&3.02e+00    &0.5448&1.76e-03&4.58e+00    &0.8913&1.75e-03&2.30e+00    &\textbf{0.9654}&\textbf{1.71e-03}&\textbf{1.97e+00} \\
 2\%   &0.3687&6.13e-03&2.95e+00    &0.5673&5.47e-03&5.37e+00    &0.7407&5.52e-03&2.51e+00    &\textbf{0.9238}&\textbf{5.31e-03}&\textbf{1.99e+00} \\
 5\%   &0.3694&1.61e-02&3.33e+00    &0.4853&1.58e-02&6.71e+00    &0.4409&1.57e-02&2.85e+00    &\textbf{0.5483}&\textbf{1.56e-02}&\textbf{2.65e+00} \\
\hline
\end{tabular}}
\vspace{-2ex}
\end{table*}

\subsection{Simulation on Synthetic Data}
Here, we generate the synthetic data as follows:
\begin{align}
\small
    \vy = \mX\vbeta + \vepsilon,
\end{align}
where $\mX\in\sR^{n\times p}$, $\vy\in\sR^n$, $\vbeta\in\sR^p$, and $\vepsilon\sim\gN(0, \alpha^{-1})$. To introduce multicollinearity to $\mX$ and $\vbeta$, we assume that
\begin{align}
\small
    \vbeta &= \mQ \vb, \\
    \mX &= \mZ \mQ^{-1},
\end{align}
where $\vb\in\sR^{p_z}$, $\mZ\in\sR^{n\times p_z}$, and $\mQ \in\sR^{p_z \times p_z}$ denotes the orthogonal part of the QR decomposition of a certain matrix $\mW$. Note that $p_z \leq p$. By specifying $\mW$ to be an identity matrix and $p_z = p$, the resulting features $\mX$ are independent of each other (or absent of multicollinearity). On the other hand, by specifying $\mW$ to be a random matrix and $p_z < p$, we introduce some multicollinearity to $\mX$ (a.k.a. partial multicollinearity). Finally, by specifying $\mW$ to be a selection matrix where each row only has one non-zero element that can be either $1$ or $-1$ and $p_z < p$, the correlation between two arbitrary columns in $\mX$ can be $1$ or $-1$. We refer to this case as ``perfect multicollinearity''. 

\begin{table*}[t]
\centering
\caption{Results of Synthetic Data with different ratios of missing values in $\mX$ averaged over 100 trials. Here, $n=100$, $p=1000$, $\alpha^{-1}= 0.01$, and the proportion of nonzero coefficients$=0.5\%$.} 
\vspace{-2ex}
\label{tab:syn_missing_values}
\resizebox{\textwidth}{!}{
\begin{tabular}{c|ccc|ccc|ccc|ccc}
\hline
 \multirow{2}*{\shortstack[l]{missing\\values}}  & \multicolumn{3}{c|}{\bf Lasso} & \multicolumn{3}{c|}{\bf E-Net} & \multicolumn{3}{c|}{\bf ARD} & \multicolumn{3}{c}{\bf BMFS}\\
\cline{2-13}
 & $F_1$-score    & MSE   & Time (s)  & $F_1$-score    & MSE   & Time (s) & $F_1$-score    & MSE   & Time (s)  & $F_1$-score    & MSE   & Time (s)\\
\hline
10\%    &0.4306&\textbf{6.65e-04}&3.72e+00    &0.3791&6.87e-04&3.98e+00    &0.7461&7.92e-04&4.23e+00    &\textbf{0.7494}&7.92e-04&\textbf{3.60e+00} \\
20\%    &0.4076&\textbf{7.46e-04}&\textbf{3.63e+00}    &0.4226&7.84e-04&3.98e+00    &0.7074&8.90e-04&5.47e+00    &\textbf{0.7835}&8.04e-04&4.65e+00 \\
 30\%   &0.4083&6.97e-04&\textbf{3.45e+00}    &0.4381&7.99e-04&3.93e+00    &0.6721&7.90e-04&6.80e+00    &\textbf{0.7714}&\textbf{6.78e-04}&5.94e+00 \\
 40\%   &0.4103&8.41e-04&\textbf{3.64e+00}    &0.3983&7.69e-04&4.41e+00    &0.6726&8.89e-04&9.81e+00    &\textbf{0.7670}&\textbf{7.69e-04}&8.64e+00 \\
 50\%   &0.3474&9.73e-04&\textbf{4.01e+00}    &0.3327&1.06e-03&4.97e+00    &0.6170&9.17e-04&1.40e+01    &\textbf{0.6727}&\textbf{8.68e-04}&1.30e+01 \\
\hline

\end{tabular}}
\vspace{-1ex}
\end{table*}

We investigate the impact of the ratio between $p$ and $n$, the level of multicollinearity, the level of noise, the level of sparsity, and the proportion of missing data on the performance of the proposed BALANCE method. As mentioned before, we compare BMFS with Lasso, E-Net, and ARD. Specifically for Lasso, the candidate set of the tuning parameter is $30$ values spaced evenly in the interval $[-2, 2]$ in the log scale. \yh{For E-Net, the common penalty parameter in front of both the $\ell_1$ and $\ell_2$ norm is selected from $10$ evenly spaced values in $[-2, 2]$ in the log scale, and the $\ell_1$ ratio parameter is chosen from $[0.1, 0.5, 0.9]$. The optimal tuning parameters are selected via cross validation. Since these two methods cannot deal with missing values explicitly, we replace the missing values with the mean of the corresponding candidate.} We consider three criteria, namely, $F_1$-score w.r.t. the zero pattern between the estimated and true $\vbeta$, mean squared error (MSE) between estimated and true $\vbeta$, and running time. Note that $F_1$-score is the harmonic mean of precision and recall, where precision is defined as the ratio between the number of true root causes identified by the model and the number of all root causes given by the model, and recall is defined as the ratio between the number of true root causes identified by the model and the number of true root causes. The results are summarized in Tables~\ref{tab:syn_multicollinar_dim}-\ref{tab:syn_missing_values}. To highlight the merit of BMFS in terms of zero pattern recovery, we further plot out the $F_1$-score of all methods as a function of $p$ in Figure~\ref{fig:syn_mutlicolinear_dim}.

\yh{Two major trends can be gleaned from the tables. First, BMFS typically achieves comparable or better performance than the SOTA methods in terms of both the estimation accuracy and the running time, especially when the dimension $p$ is high, and the level of multicollinearity is high. On the other hand, the superiority of BMFS and ARD over E-Net and Lasso suggests that adaptively learning the tuning parameters from the data via variational inference is more advantageous than estimating them via brute-force grid search. Grid search restricts the tuning parameters to the predefined set of candidates, and consequently, a carelessly designed set may lead to unsatisfactory performance. In addition, it can be observed that the performance of BMFS and E-Net is better than that of ARD and Lasso respectively, when there exists multicollinearity in the data, as expected. Indeed, the $\ell_1$-norm penalized Lasso can only pick at most $n$ candidates when $p >n$ in theory, even if all candidates are relevant. More precisely, if there are two or more highly collinear candidates, Lasso only selects one of them at random. This explains the deficiency of ARD and Lasso in comparison with BMFS and E-Net.}


\yh{Second, the performance of BMFS is robust to the number of dimensions, the level of multicollinearity, the noise level, the sparsity level, and the proportion of missing values. To be specific, we can see that Bayesian methods constantly outperform the frequency methods for both $p \leq n$ and $p > n$. Additionally, Bayesian methods offer a straightforward way to cope with missing values by inferring their distributions from the data at the expense of consuming more time with the increase of the missing data proportion. In a contrast, we employ the mean imputation method before applying the frequentist methods, leading to inaccurate estimation when the proportion of missing values is large.}


\subsection{Bad SQL Localization}

\begin{table}[t]
\caption{Different tenant KPIs and the corresponding SQL metrics.}
\centering
\vspace{-2ex}
\resizebox{.7\columnwidth}{!}{
\begin{tabular}{|c|c|}
\hline
\textbf{Tenant KPI}              & \textbf{SQL Metric} \\ \hline
\textit{SQL\_SELECT\_RT}                  & \textit{cpu\_time} \\ \hline
\textit{LOGICAL\_READS}                   & \textit{lr(logical\_reads)}  \\ \hline
\textit{SQL\_QUEUE\_TIME}                 & \textit{queue\_time}   \\ \hline
\textit{RPC\_PACKAGE\_IN/OUT}           & \textit{rpc\_count}      \\ \hline
\end{tabular}
}
\vspace{-2.8ex}
\label{tab:tenant_sql_metric}
\end{table}
Database services are a fundamental infrastructure that is critical for the everyday business of enterprises. Thus, it is of top priority to guarantee the high availability of database services. Previous works, such as iSQUAD~\cite{ma2020diagnosing}, concentrate on determining the fault type of an intermittent slow SQL from typical types given by experts or from historical data. \yh{Different from slow queries, which appear in massive numbers in the slow query logs, bad SQL localization is a more comprehensive problem as we consider not only run time but also logical reads, RPC count, etc. Bad SQLs cannot be always found in the slow query logs since they may lead to anomalies in the tenant KPIs via their CPU or memory usage instead of run time. As mentioned in the introduction, here we aim to find "Bad SQLs" (i.e., the candidate root causes $\mX$) that are suspicious and responsible for the anomalies detected in tenant KPIs (i.e., the target KPIs $\vy$).}

\begin{figure}[t]
  \centering
  \includegraphics[width=0.98\linewidth]{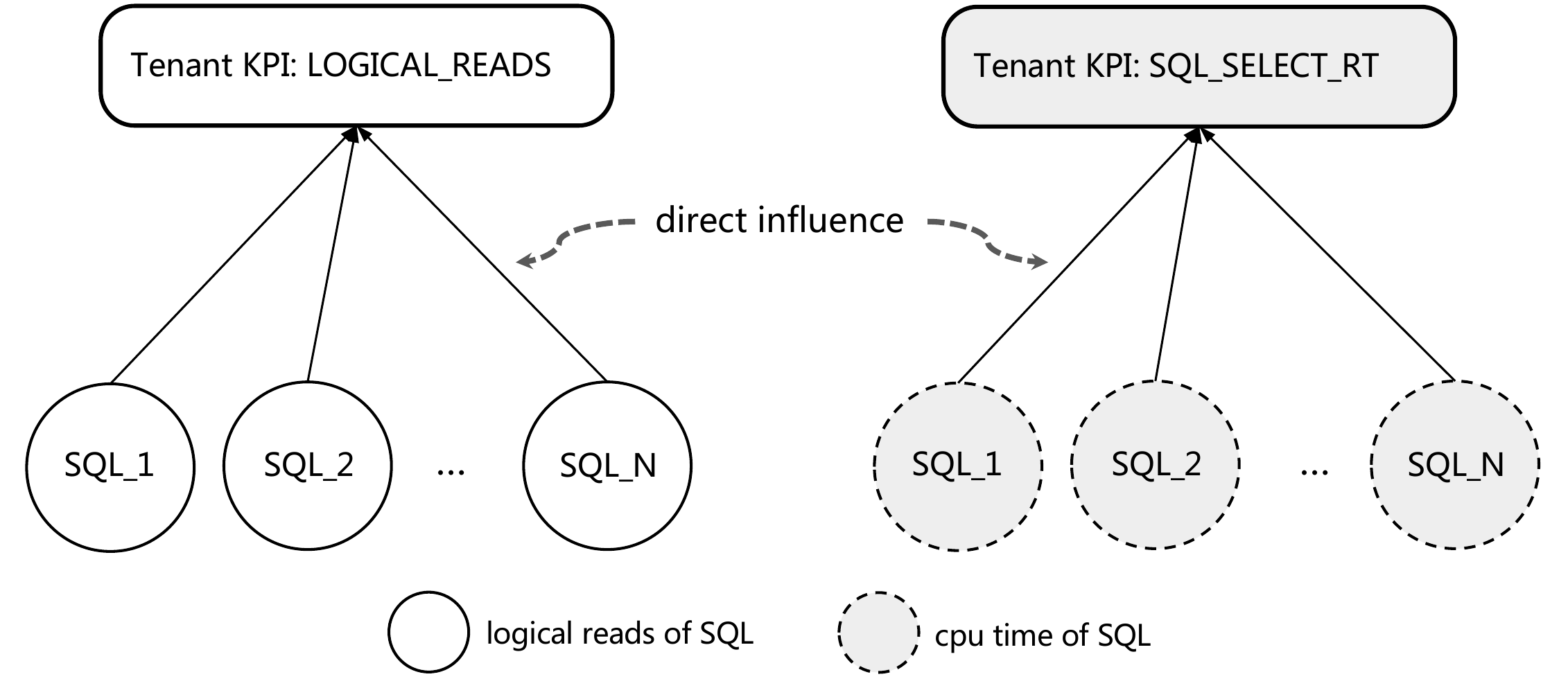}
  \vspace{-2ex}
  \caption{Targets and candidates for bad SQL localization.}
  \label{fig:sql-proc}
  \vspace{-3.5ex}
\end{figure}

As shown in Table~\ref{tab:tenant_sql_metric}, Tenant KPIs monitor the performance of tenants, while SQL metrics tell us the performance of each SQL. In light of expert knowledge and offline data analysis, we find that almost all bad SQL issues can be reflected by two kinds of tenant KPIs, that is, \textit{SQL\_SELECT\_RT} and \textit{LOGIC\_READS}. As such, the alarm of these two KPIs is the trigger of our RCA module. Moreover, we find that the tenant KPIs can be regarded as a summary of the relevant SQL metrics. For instance, the tenant KPI \textit{SQL\_SELECT\_RT} is influenced by the metric \textit{cpu\_time} of all SQLs, while the other KPI \textit{LOGIC\_READS} is associated with the metric \textit{lr} (i.e., logic reads). Hence, this application of BALANCE considers homogeneous $\mX$ and $\vy$. Figure~\ref{fig:sql-proc} displays the targets and candidate pairs to be analyzed by BALANCE in bad SQL localization.

In addition, on the online monitor platform, the tenant KPIs are saved as fixed-interval time series, whereas the SQL metrics are collected by batch sampling: each batch collects the metrics of all SQLs within 10 seconds before the collection time, and the next collection starts after the previous one finishes the storage process and so there are small uneven time gaps between every two collections that amount to the storage time. To settle this problem, we aggregate the SQL metrics within every minute and regard them as time series. As a result, the timestamps of $\mX$ and $\vy$ might not be aligned. We, therefore, use both $\mX^t$ and $\mX^{t-1}$ when aligning with $\vy^t$, and output the SQL $x_j$ as a possible bad SQL when the coefficient in front of either $\vx_j^t$ or $\vx_j^{t-1}$ is non-zero. 

\yh{Finally, while there are only two target KPIs, the number of related SQLs $p$ can be very large (e.g., thousands or larger), varying in each case.} The SQLs are often correlated with each other as a small modification of the WHERE or the LIMIT condition in a SQL sentence is defined as a new SQL. The proportion of missing data can also be very large. Furthermore, it is required to find the root causes within 1 minute, and the results should be interpretable. In a nutshell, under the problem of bad SQL localization, all 6 desiderata in $
\S$\ref{sec:desiderata} should be satisfied. To merge the root causes resulting from the two KPIs, we retain the top 3 root causes for each KPI (i.e., $\kappa = 3$ in the merging module), and compute the union of the two sets of root causes as the final recommendation.

To evaluate the performance of the proposed approach, we collect 90 samples from the SQL diagnosing platform of Ocean Base\cite{yangxu2022OB}. The samples cover over 50 different tenants, and \yh{in each sample, both the tenant KPIs and the SQL metrics are saved as 61-minute time series (i.e., $n = 61$), including the one-minute abnormal part. Recall that $p$ scales up to hundreds or thousands. Hence, $p > n$ in this scenario.} Missing data typically exist in the SQL metrics, up to $50\%$. To match the requirement of real-time RCA tasks, three evaluation criteria are selected that is accuracy, number of recommendations, and running time. For each sample, if the true root causes lie within the set of SQLs recommended by the proposed method, we count this sample as a ``hit''; otherwise, it is a ``miss''. Accuracy is then defined as the proportion of the number of hits to the overall number of samples. Again, we compare different forward models. As Lasso and E-Net cannot deal with missing data, we impute the missing values using linear interpolation for tenant KPIs and fill zeros for the SQL metrics. \yh{In addition, we further consider another benchmark method, fsMTS (feature selection for multivariate time series)~\cite{pavlyuk2022fsmts}. This method considers lagged temporal dependence and selects features from $\mX_{t}, \cdots, \mX_{t-\tau}$, where $\tau$ is the time lag. It then resorts to an ensemble method to combine the results from cross-correlation, graphical lasso, random forest, etc. As fsMTS can only tell us whether a feature is selected, we choose the tuning parameter such that only $\kappa = 3$ candidates are selected for each KPI.} The results for all methods are presented in Table~\ref{tab:bad_sql_accuracy_efficiency}. 

\begin{table}[]
\centering
\caption{Results for bad SQL localization.}
\vspace{-2ex}
\resizebox{\columnwidth}{!}{
\begin{tabular}{|c|c|c|c|c|c|}
\hline
\textbf{Methods}       & \#Hits & \#Misses & \textbf{Accuracy} & \textbf{Time (s)}& \textbf{\#Recommend} \\ \hline
ARD                 & 68     & 24     & 0.7556           & 1.92e+00   & 2.7       \\ \hline
Lasso               & 64     & 24     & 0.7111         & 3.15e+00   & 2.1        \\ \hline
E-Net                & 64     & 26     & 0.7222           & 2.95e+00   & 2.3        \\ \hline
fsMTS                & 49     & 41     & 0.5444          & 1.20e+01    & 5.0      \\ \hline
BMFS                & 75     & 15     & 0.8333          & 1.78e+00    & 2.3      \\ \hline
\end{tabular}
}
\label{tab:bad_sql_accuracy_efficiency}
\vspace{-3ex}
\end{table}

It can be observed that the proposed method achieves the highest accuracy with the lowest number of recommendations and the lowest amount of computational time. The accuracy is $83.33\%$, which is at least $8\%$ higher than the baseline methods. The number of recommendations is only $2.3$, and the running time is only $1.78$ seconds on average. Note that while being complete, the number of recommendations is supposed to be as small as possible as it helps the SREs to concentrate on a few root causes so they can dive in and fix the issue more efficiently. On the other hand, we can tell from Table~\ref{tab:bad_sql_accuracy_efficiency} that the performance of BMFS and E-Net is better than that of ARD and Lasso respectively since the former two methods can cope with the multicollinearity within the data. We can also find that the Bayesian methods (i.e., BMFS and ARD) outperform their frequency counterparts (i.e., E-Net and Lasso) since they can learn the distribution of the tuning parameters as well as the missing values adaptively from the data. We have deployed the proposed method to production, and the online results show that the accuracy can be as high as $95\%$, which further demonstrates the advantages of BALANCE. \yh{Finally, fsMTS yields the worst result, probably because 1) temporal dependence is not the main concern for time series grouped by the minute; 2) the base learner in the ensemble, such as graphical lasso, also needs careful tuning of the penalty parameters but now only the default value is used.}

\vspace{-1ex}
\subsection{Container Fault Localization}
In this section, we further apply the proposed approach to another practical situation, that is, container fault localization. In this situation, once the number of trace failures associated with a container is abnormal, the proposed BALANCE method would find the container metrics that can best explain the anomaly, and the RCA results will further lead to self-healing operations such as restart and traffic throttling. By addressing all trace failures automatically, the high availability of the cloud-native system can be guaranteed. 

\begin{figure*}[]
    \centering
    \subfigure[Target KPI $\vy$]{
    \includegraphics[width = 0.242\linewidth]{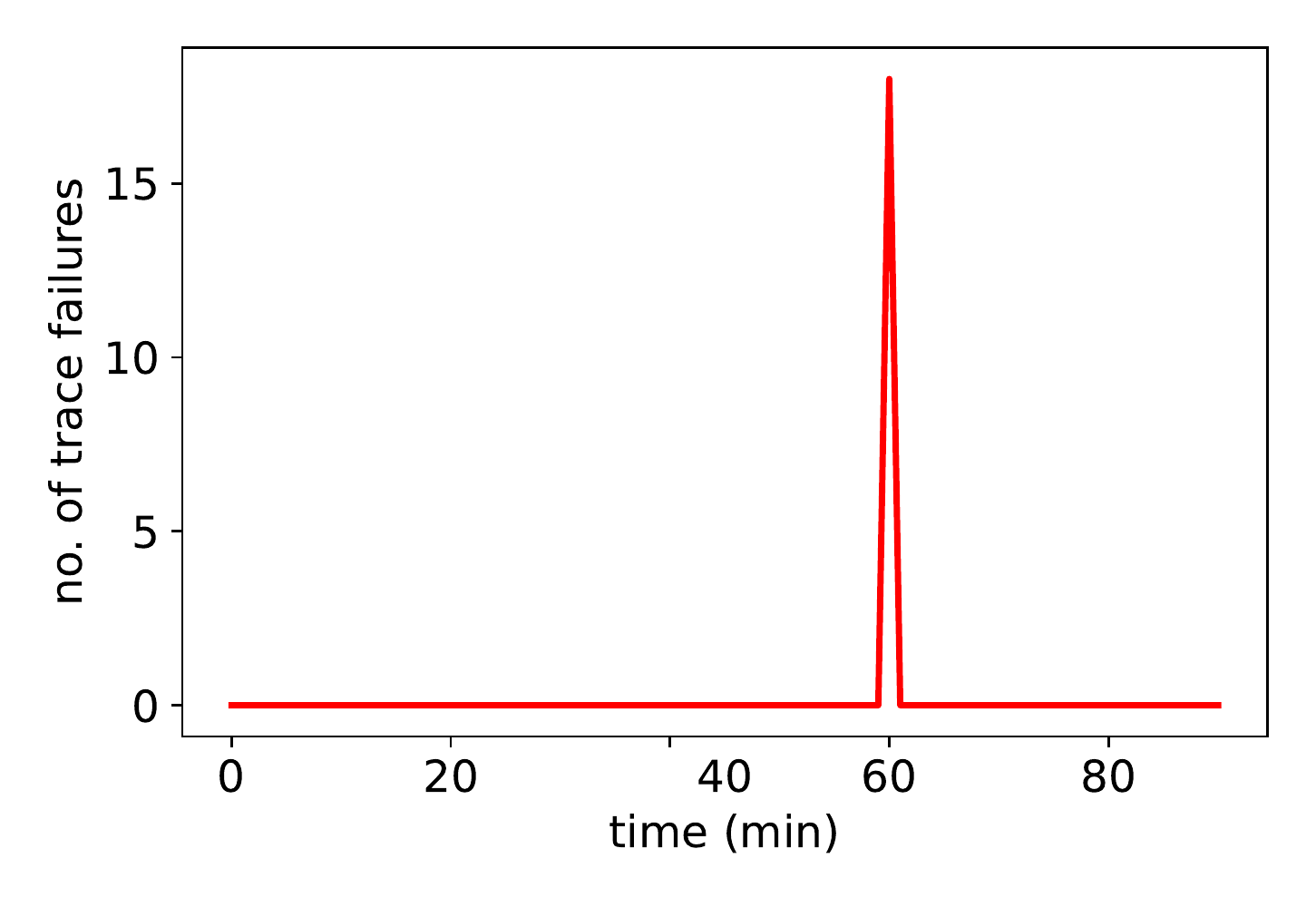}
    \label{sfig:trace_failures}
    }
    \hspace{-1em}
    \subfigure[CPU related metrics]{
    \includegraphics[width = 0.242\linewidth]{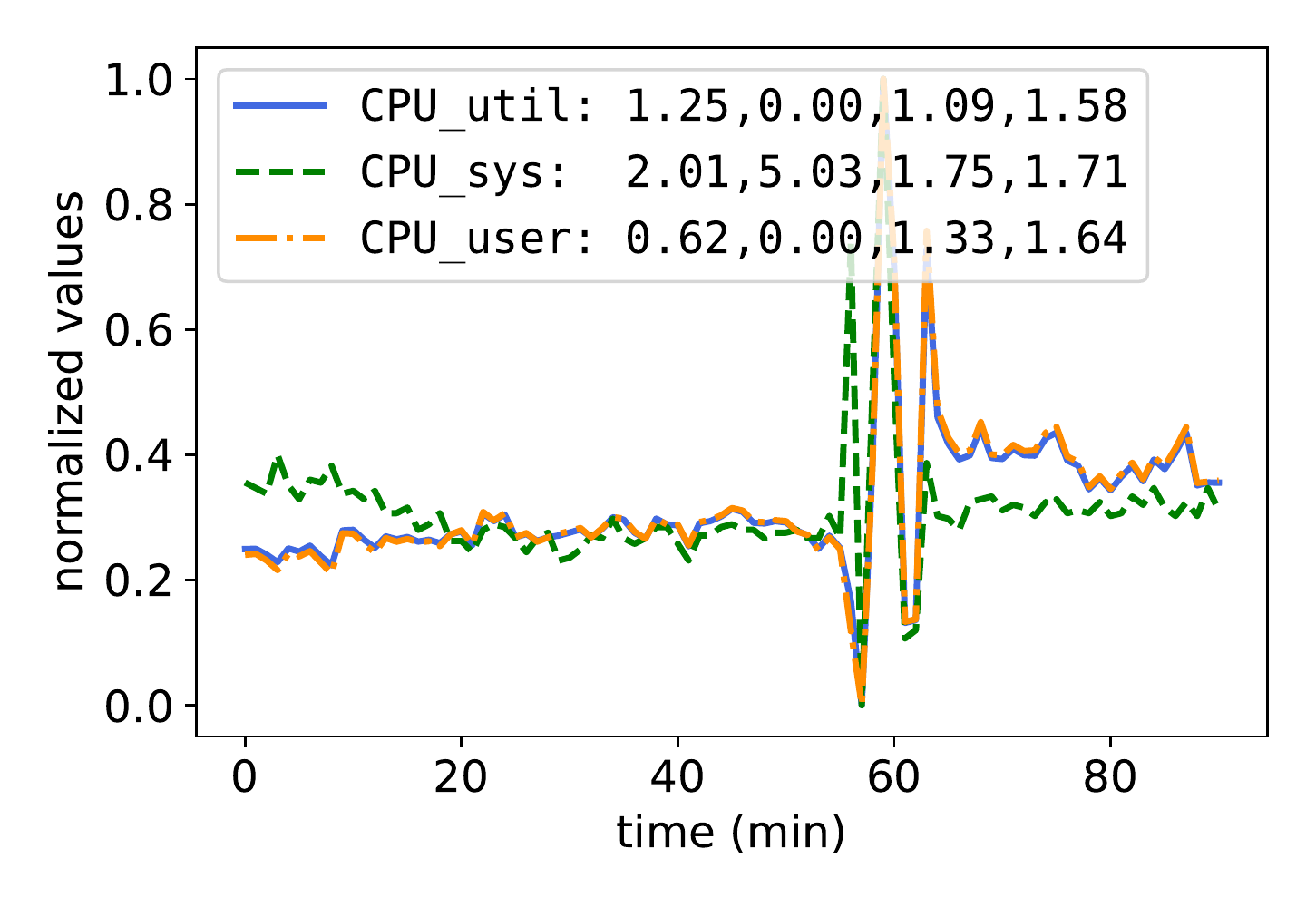}
    \label{sfig:cpu_metrics}
    }
     \hspace{-1em}
    \subfigure[Memory related metrics]{
    \includegraphics[width = 0.242\linewidth]{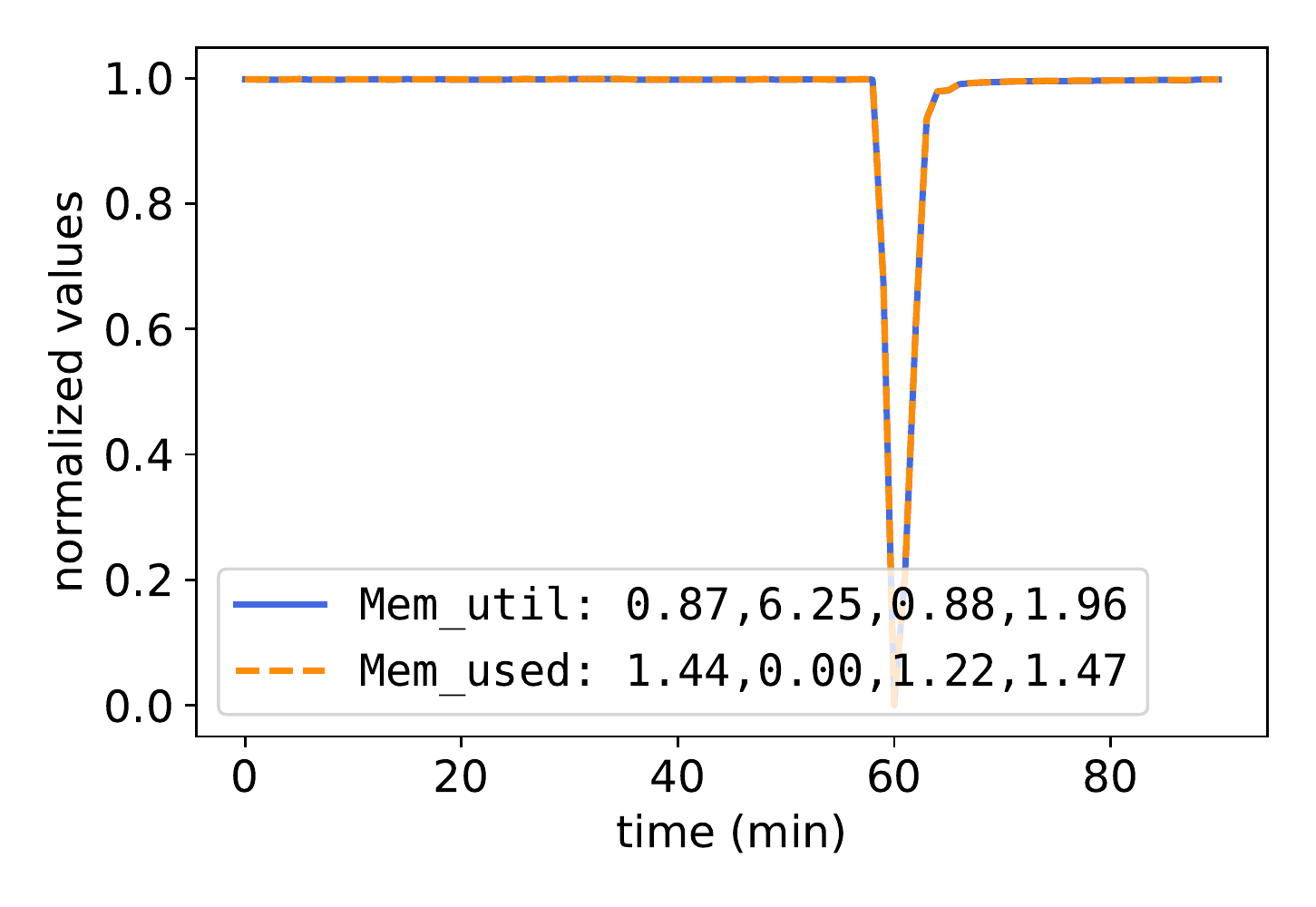}
    \label{sfig:mem_metrics}
    }
     \hspace{-1em}
    \subfigure[Traffic related metrics]{
    \includegraphics[width = 0.242\linewidth]{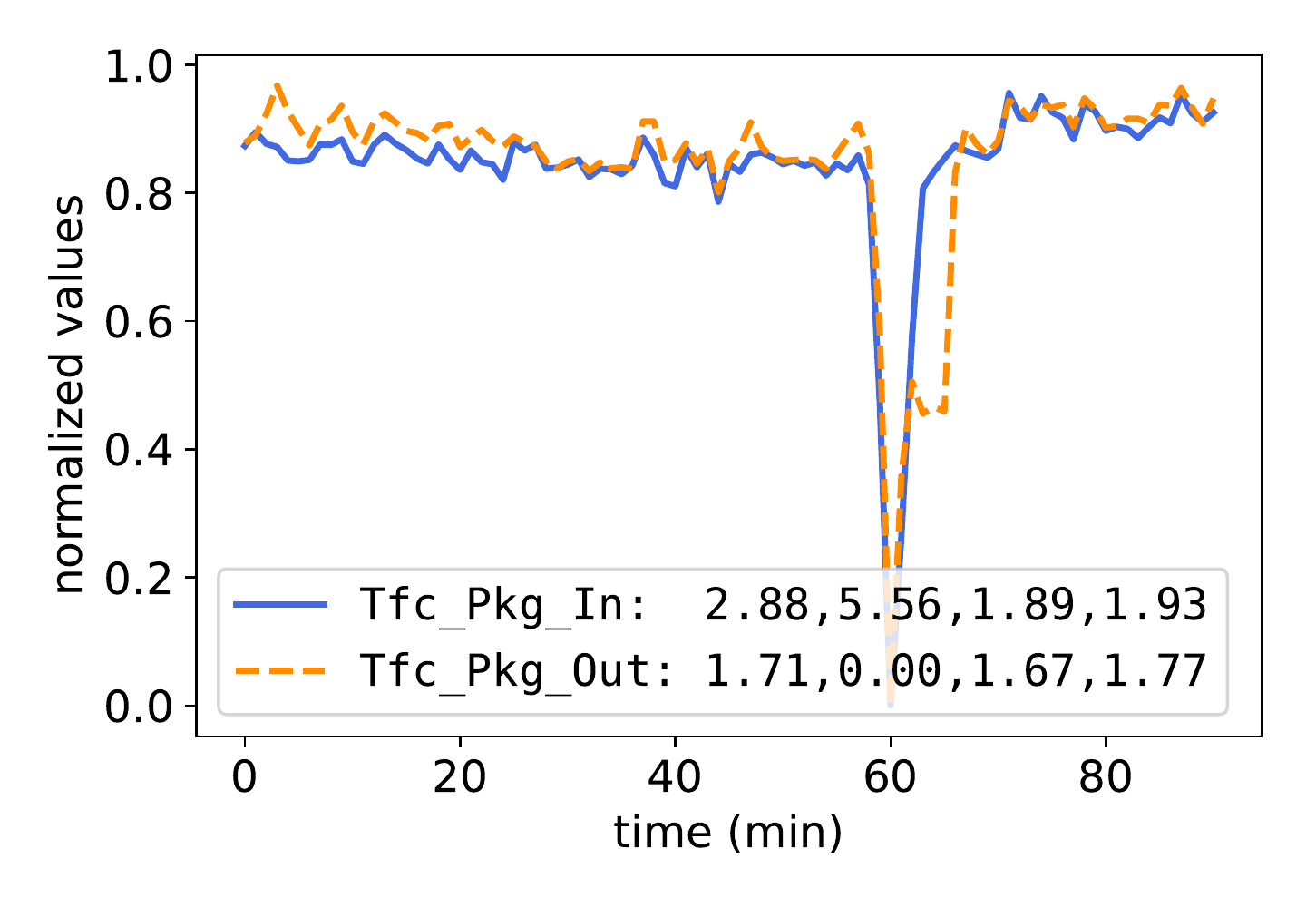}
    \label{sfig:traffic_metrics}
    }
    \vspace{-3ex}
  \caption{Visualization of the trace failure number $\vy$ and correlated container metrics: the four values associated with each metric in the legend denote the attribution scores of ARD, Lasso, E-Net, and BMFS sequentially.}
  \label{fig:multicollinear_examples}
  \vspace{-2ex}
\end{figure*}

\yh{In this case, the target KPI $\vy$ is the number of trace failures associated with a container, and the candidate root causes $\mX$ is the $10$ metrics of this container}, such as CPU usage, memory usage, inward and outward traffic, the number of TCP connections, etc. Note that $\vy$ and $\mX$ are heterogeneous under this setting. Moreover, there exist strong correlations among the container metrics. For instance, the metric CPU usage is the sum of another two metrics: CPU user usage and CPU system usage, and additionally, the metrics inward traffic in bytes, outward traffic in bytes, inward traffic in packages, and outward traffic in packages are closely correlated. Missing data also appears now and then, up to $10\%$. 

To assess the quality of our method, we collect 100 samples in total from a fault diagnosing platform of Ant Group. For each sample, we can obtain the alarm time and 30-minute time series for both $\mX$ and $\vy$ before and during the anomaly. \yh{In this case, $n = 30$, $p = 10$, and $p < n$.} We benchmark the proposed forward model BMFS against Lasso, E-Net, ARD, and fsMTS, and evaluate all models in terms of precision, recall, $F_1$-score, and running time. The results are listed in Table~\ref{tab:Table 5}. Notice that the deployment of BALANCE to production yields similar results. BMFS blows out of the water Lasso, E-Net, ARD, and fsMTS in terms of both $F_1$-score and running time. Indeed, the $F_1$ score resulting from BMFS is above $20\%$ higher than that of the second best method ARD. \yh{Concretely, the precision of BMFS and E-Net is respectively similar to that of ARD and Lasso, but the recall of the former two methods is much higher, indicating that ARD and Lasso mistakenly omit some relevant metrics and highlighting the appeal of considering multicollinearity among candidate root causes. To further elucidate this point, we depict as an example the target KPI and the correlated candidate root causes in Figure~\ref{fig:multicollinear_examples} and show the attribution scores for ARD, Lasso, E-Net, and BMFS sequentially for each candidate root cause in the figure. It can be seen that BMFS yields similar attribution scores for closely correlated root causes (i.e., the axiom of continuity in $\S$\ref{ssec:attribution}), whereas the attribution scores resulting from other methods do not always follow this axiom.} In addition, we can observe that the $F_1$-score of both BMFS and ARD is higher than that of E-Net and Lasso, implying the merits of learning the distributions of tuning parameters adaptively from the data. Finally, despite the heterogeneity between $\mX$ and $\vy$ in this scenario, the proposed BALANCE still works well, probably due to the model-agnostic nature of linear attribution methods~\cite{lundberg2017unified}. On the other hand, we notice that the relationship between heterogeneous $\vx_j$ and $\vy$ during anomalies can be well approximated by a linear model since similar phenomena (i.e., the waveform anomaly) usually occurs to both of them during anomalies. For example, an abnormal spike in $\vx_j$ typically leads to a spike in $\vy$, while an abrupt increase in $\vx_j$ results in an increase or decrease in $\vy$.

\begin{table}[t]
\caption{Results for container fault localization.}
\vspace{-2ex}
\centering
\resizebox{.75\columnwidth}{!}{
\begin{tabular}{|c|c|c|c|c|}
\hline
\textbf{Methods}       & Recall & Precision &$F_1$-score & \textbf{Time (s)}\\ \hline
ARD & 0.4784     & 0.98804    & 0.6266           & 3.54e-1        \\ \hline
Lasso & 0.5083     & 0.5219     & 0.4712           & 4.96e-1          \\ \hline
E-Net  & 0.7352     & 0.5520     & 0.5925           & 4.58e-1          \\ \hline
fsMTS    & 0.4826     & 0.5244     & 0.4655           & 7.01e-1         \\ \hline
BMFS     & 0.8180     & 0.9136     & 0.8569           & 1.52e-1         \\ \hline
\end{tabular}
}
\vspace{-3ex}
\label{tab:Table 5}
\end{table}

\vspace{-1ex}
\subsection{\yh{Fault Type Diagnosis for Exathlon Dataset}}
\yh{We further investigate the performance of BALANCE on the public benchmark, Exathlon~\cite{jacob2021exathlon}, which consists of real data traces collected from 10 Spark streaming applications in a 4-node cluster over a 2.5-month period. There are 5 types of faults injected into the traces during executions, including Bursty Input (Type 1), Bursty Input Until Crash (Type 2), Stalled Input (Type 3), CPU Contention (Type 4), and Process Failure (Type 5). The duration of the fault injection approximately ranges from 15 to 30 minutes. 
For each injection, the time series of 2,283 metrics coming from Spark are collected. Originally, Exathlon conducts anomaly detection for multivariate time series and then attributes the anomaly of multivariate time series as a whole to individual time series. Here, we build a new fault-type diagnosis task on top of Exathlon (named Exathlon-ftd) to evaluate our method. Exathlon-ftd is constructed in the following way.}

\yh{\textit{First}, we define the target KPI $\vy$ and the candidate metrics $\mX$ for this dataset. The target KPI is supposed to be affected by various types of faults and can be used to trigger the RCA procedure. 
To this end, we conduct anomaly detection using the $3\sigma$ rule after detrending on all metrics for each injection and find the intersection of the anomalous metrics in all cases. We choose the metric \textit{Processing\_Delay} from the intersection as our target KPI $\vy$, since the processing time is influenced by all aforementioned types of faults both theoretically and empirically. The candidate metrics $\mX$ is then chosen as the metrics that are abnormal in the same duration when $\vy$ is abnormal via the candidate AD module in Figure~\ref{fig:larc-framework}. }

\yh{\textit{Second}, we establish rules 
to determine the fault types once the root causes metrics are selected from $\mX$. For simplicity, we only focus on the first 4 types of events and define 4 rules (R1-R4) to determine the fault types given the estimated root cause metrics as:
\begin{itemize}[leftmargin=2em,itemsep= 2pt,topsep = 2pt,partopsep=2pt]
\item[\textit{R1}.]\label{rule1} If the metrics about traffic such as \textit{Last\_Complete\_Batch\_Records} and \textit{Processed\_Records} are in the estimated set of root causes, then it is the problem of the input (i.e., traffic). The fault type is ``Bursty Input'' or ``Bursty Input Until Crash'' if these metrics are increasing and is ``Stalled Input'' if decreasing.
\item[\textit{R2}.]\label{rule2} If \textit{R1} is not satisfied and the estimated set of root causes contains the metrics about CPU that are concentrated on a single node, then the fault type is ``CPU contention''. The rationale behind is that the traffic problem can lead to anomalies in the metrics about CPU, but the CPU contention problem cannot in turn influence the metrics about traffic.  
\item[\textit{R3}.]\label{rule3} If the above two rules are not satisfied and the estimated set of root causes contains the metric \emph{executor\_active\_tasks}, then the fault type is ``Bursty Input'' or ``Bursty Input Until Crash''. We notice that both the traffic increase and the CPU contention problem will cause the abnormal increase in \emph{executor\_active\_tasks}. However, a CPU contention problem typically results from another program on a single node that consumes all the CPU cores available on that node, and thus, the root cause metrics about CPU in the estimated set typically resolve around a single node (i.e., \textit{R2}). As this condition does not hold, we can conclude that the fault type here is ``Bursty Input''.
\item[\textit{R4}.]\label{rule4} If the above three rules are not satisfied, the fault type is ``Unknown''.
\end{itemize}
}

\yh{
After these two steps, 73 cases are collected in total for the RCA task. For each case, we aggregate both $\mX$ and $\vy$ by minute, and we retain 60 minutes and 5 minutes respectively before and after \emph{Root\_Cause\_State} time available in the dataset. 
Thus, $n=66$. The number of candidates $\mX$ after anomaly detection scales up to $p = 150$ approximately. We then compare all methods on the constructed Exathlon-ftd task. The results are presented in Table~\ref{tab:exathlon_accuracy_efficiency}. Once again, BMFS outperforms all the compared methods with the highest accuracy and the fewest number of recommended root cause metrics. The running time of BMFS is the second best, slightly longer than that of ARD.
}

\begin{table}[t]
\centering
\caption{\yh{Results for Exathlon fault type analysis.}}
\vspace{-1ex}
\yh{
\resizebox{\columnwidth}{!}{
\begin{tabular}{|c|c|c|c|c|c|}
\hline
\textbf{Methods}    & \#Hits    & \#Misses  & \bf{Accuracy}     & \bf{Time (s)}     & \bf{\#Recommend}      \\ \hline
ARD	                                        & 54	    & 19	    & 0.7397	        & 1.29e+00	        & 3.4                   \\ \hline
Lasso                                       & 51        & 22        & 0.6986            & 1.58e+00          & 4.4                   \\ \hline
E-Net                                       & 49        & 24        & 0.6712            & 1.67e+00          & 4.7                   \\ \hline
fsMTS                                       & 36        & 37        & 0.4932            & 4.45e+01          & 4.9                   \\ \hline
BMFS	                                    & 58	    & 15	    & \bf{0.7945}	    & 1.60e+00	        & 3.2                   \\ \hline
\end{tabular}
}
}
\vspace{-3ex}
\label{tab:exathlon_accuracy_efficiency}
\end{table}

\section{Conclusion and Future Work}
The major contribution of this paper is to shed a different light on the RCA problem, viewing it from the angle of XAI. In particular, we assume that the abnormal behavior of the target KPIs can be explained by the relevant candidate root causes and propose a novel attribution-based RCA method named BALANCE. Concretely, we first learn a Bayesian sparse multicollinear model that predicts the target KPIs given the candidate root causes. We then attribute the abnormal behavior of the target KPIs to the candidates by computing their attribution scores. We empirically show that the proposed method achieves superior performance for the synthetic data and three real-world problems, including bad SQL localization, container fault localization, and fault type diagnosis. We notice that the target KPIs and the candidate root causes are homogeneous in the former case and heterogeneous in the latter two cases, \yh{while $p > n$ in the first and third application and $p \leq n$ in the second one.} We have deployed BALANCE to production, providing real-time root cause diagnosis of bad SQL and container issues in distributed data systems. Although far from exhaustive, these applications show that BALANCE has the potential to serve as a general tool for practical RCA tasks.

In future work, we would like to explore the connections between attribution-based RCA and graph-based RCA. 
In addition, we would like to generalize the linear model in 
BALANCE to non-linear forward models, such as the generalized linear models, which can be more expressive and suitable for describing the relationship between heterogeneous metrics.

\bibliographystyle{ACM-Reference-Format}
\bibliography{BALANCE}
\end{document}